\documentclass[12pt]{article}
\usepackage[blocks]{authblk}
\usepackage[tt=false, type1=true]{libertine}

\usepackage{amsmath}
\usepackage{amsthm}
\usepackage{amssymb}
\usepackage{amsfonts}
\usepackage{mathrsfs}  
\usepackage{hyperref}
\usepackage{url}
\usepackage{graphicx}
\usepackage{array}
\usepackage{booktabs}
\usepackage{xypic}
\usepackage{algorithm}
\usepackage{subcaption}
\usepackage{multirow}
\usepackage{setspace}
\usepackage[noend]{algpseudocode}

\usepackage{soul}

\algrenewcommand\algorithmiccomment[1]{\hfill \# \textit{#1}}

\usepackage[tikz]{bclogo}

\newtheorem{thm}{Theorem}
\newtheorem{lem}{Lemma}
\newtheorem{defn}{Definition}

\DeclareMathOperator{\Real}{\mathsf{Real}}
\DeclareMathOperator{\Sing}{\mathsf{Sing}}
\DeclareMathOperator{\FinReal}{\mathsf{FinReal}}
\DeclareMathOperator{\FinSing}{\mathsf{FinSing}}
\DeclareMathOperator*{\colim}{colim}

\newcolumntype{P}[1]{>{\centering\arraybackslash}p{#1}}

\title{\vspace{-2.5cm}\textsf{\bfseries UMAP: Uniform Manifold Approximation and Projection for Dimension Reduction}}

\author{Leland McInnes}
\affil{Tutte Institute for Mathematics and Computing\\ leland.mcinnes@gmail.com}
\author{John Healy}
\affil{Tutte Institute for Mathematics and Computing\\ jchealy@gmail.com}
\author{James Melville}
\affil{jlmelville@gmail.com}

\begin{document}

\maketitle

\abstract{
UMAP (Uniform Manifold Approximation and Projection) is a novel manifold learning technique for dimension reduction.  UMAP is constructed from a theoretical framework based in Riemannian geometry and algebraic topology.  The result is a practical scalable algorithm that is applicable to real world data.  The UMAP algorithm is competitive with t-SNE for visualization quality, and arguably preserves more of the global structure with superior run time performance.  Furthermore, UMAP has no computational restrictions on embedding dimension, making it viable as a general purpose dimension reduction technique for machine learning.
}

\section{Introduction}

Dimension reduction plays an important role in data science, being a fundamental technique in both visualisation and as pre-processing for machine learning. Dimension reduction techniques are being applied in a broadening range of fields and on ever increasing sizes of datasets.  It is thus desirable to have an algorithm that is both scalable to massive data and able to cope with the diversity of data available. Dimension reduction algorithms tend to fall into two categories; those that seek to preserve the pairwise distance structure amongst all the data samples and those that favor the preservation of local distances over global distance.  Algorithms such as PCA \cite{hotelling1933analysis}, MDS \cite{Kruskal1964}, and Sammon mapping \cite{sammon1969nonlinear} fall into the former category while t-SNE \cite{maaten2008visualizing,van2014accelerating}, Isomap \cite{tenenbaum2000global}, LargeVis \cite{tang2016visualizing}, Laplacian eigenmaps \cite{belkin2002laplacian,belkin2003laplacian} and diffusion maps \cite{coifman2006diffusion} all fall into the latter category.  

In this paper we introduce a novel manifold learning technique for dimension reduction.  We provide a sound mathematical theory grounding the technique and a practical scalable algorithm that applies to real world data. UMAP (Uniform Manifold Approximation and Projection) builds upon mathematical foundations related to the work of Belkin and Niyogi on Laplacian eigenmaps.  We seek to address the issue of uniform data distributions on manifolds through a combination of Riemannian geometry and the work of David Spivak \cite{spivakmetric} in category theoretic approaches to geometric realization of fuzzy simplicial sets.  t-SNE is the current state-of-the-art for dimension reduction for visualization.  Our algorithm is competitive with t-SNE for visualization quality and arguably preserves more of the global structure with superior run time performance. 
Furthermore the algorithm is able to scale to significantly larger data set sizes than are feasible for t-SNE.
Finally, UMAP has no computational restrictions on embedding dimension, making it viable as a general purpose dimension reduction technique for machine learning.

Based upon preliminary releases of a software implementation, UMAP has already found widespread use in the fields of bioinformatics \cite{becht2019dimensionality, cao2019single, diaz2018revealing, park2018fast, bagger2018bloodspot, Oetjen416750, clark2018comprehensive}, materials science \cite{li2019manifold, fuhrimann2018data}, and machine learning \cite{carter2019activation, espadoto2019deep, espadotovisual, gaujac2018gaussian, escolano2018self, posada2018msc} among others. 

This paper is laid out as follows. In Section \ref{theory} we describe the theory underlying the algorithm.  Section \ref{theory} is necessary to understand both the theory underlying why UMAP works and the motivation for the choices that where made in developing the algorithm.  A reader without a background (or interest) in topological data analysis, category theory or the theoretical underpinnings of UMAP should skip over this section and proceed directly to Section \ref{computational}.

That being said, we feel that strong theory and mathematically justified algorithmic decisions are of particular importance in the field of unsupervised learning.  This is, at least partially, due to plethora of proposed objective functions within the area. We attempt to highlight in this paper that UMAPs design decisions were all grounded in a solid theoretic foundation and not derived through experimentation with any particular task focused objective function.  Though all neighbourhood based manifold learning algorithms must share certain fundamental components we believe it to be advantageous for these components to be selected through well grounded theoretical decisions. One of the primary contributions of this paper is to reframe the problem of manifold learning and dimension reduction in a different mathematical language allowing pracitioners to apply a new field of mathemtaics to the problems. 

In Section \ref{computational} we provide a more computational description of UMAP.   Section \ref{computational} should provide readers less familiar with topological data analysis with a better foundation for understanding the theory described in Section \ref{theory}.  Appendix \ref{compare} contrasts UMAP against the more familiar algorithms t-SNE and LargeVis, describing all these algorithms in similar language. This section should assist readers already familiar with those techniques to quickly gain an understanding of the UMAP algorithm though they will grant little insite into its theoretical underpinnings.

In Section \ref{implementation} we discuss implementation details of the UMAP algorithm. This includes a more detailed algorithmic description, and discussion of the hyper-parameters involved and their practical effects.

In Section \ref{experiments} we provide practical results on real world datasets as well as scaling experiments to demonstrate the algorithm's performance in real world scenarios as compared with other dimension reduction algorithms. 

In Section \ref{weaknesses} we discuss relative weakenesses of the algorithm, and applications for which UMAP may not be the best choice.

Finally, in Section \ref{future} we detail a number of potential extensions of UMAP that are made possible by its construction upon solid mathematical foundations.  These avenues for further development include semi-supervised learning, metric learning and heterogeneous data embedding.  

\section{Theoretical Foundations for UMAP}\label{theory}

The theoretical foundations for UMAP are largely based in manifold theory and topological data analysis. Much of the theory is most easily explained in the language of topology and category theory. Readers may consult \cite{mac2013categories}, \cite{riehl2017category} and \cite{may1992simplicial} for background. Readers more interested in practical computational aspects of the algorithm, and not necessarily the theoretical motivation for the computations involved, may wish to skip this section.  Readers more familiar with traditional machine learning may find the relationships between UMAP, t-SNE and Largeviz located in Appendix {\ref{compare}} enlightening.  Unfortunately, this purely computational view fails to shed any light upon the reasoning that underlies the algorithmic decisions made in UMAP.  Without strong theoretical foundations the only arguments which can be made about algorithms amount to empirical measures, for which there are no clear universal choices for unsupervised problems.

At a high level, UMAP uses local manifold approximations and patches together their local fuzzy simplicial set representations to construct a topological representation of the high dimensional data. Given some low dimensional representation of the data, a similar process can be used to construct an equivalent topological representation. UMAP then optimizes the layout of the data representation in the low dimensional space, to minimize the cross-entropy between the two topological representations.

The construction of fuzzy topological representations can be broken down into two problems: approximating a manifold on which the data is assumed to lie; and constructing a fuzzy simplicial set representation of the approximated manifold. In explaining the algorithm we will first discuss the method of approximating the manifold for the source data. Next we will discuss how to construct a fuzzy simplicial set structure from the manifold approximation. Finally, we will discuss the construction of the fuzzy simplicial set associated to a low dimensional representation (where the manifold is simply $\mathbb{R}^d$), and how to optimize the representation with respect to our objective function.

\subsection{Uniform distribution of data on a manifold and geodesic approximation}\label{graph_derivation}

The first step of our algorithm is to approximate the manifold we assume the data (approximately) lies on. The manifold may be known apriori (as simply $\mathbb{R}^n$) or may need to be inferred from the data. Suppose the manifold is not known in advance and we wish to approximate geodesic distance on it. Let the input data be $X = \{X_1, \ldots, X_N\}$. As in the work of Belkin and Niyogi on Laplacian eigenmaps \cite{belkin2002laplacian,belkin2003laplacian}, for theoretical reasons it is beneficial to assume the data is uniformly distributed on the manifold, and even if that assumption is not made (e.g \cite{hein2007graph}) results are only valid in the limit of infinite data. In practice, finite real world data is rarely so nicely behaved. However, if we assume that the manifold has a Riemannian metric not inherited from the ambient space, we can find a metric such that the data is approximately uniformly distributed with regard to that metric.

Formally, let $\mathcal{M}$ be the manifold we assume the data to lie on, and let $g$ be the Riemannian metric on $\mathcal{M}$. Thus, for each point $p\in \mathcal{M}$ we have $g_p$, an inner product on the tangent space $T_p\mathcal{M}$.

\begin{lem}\label{lem:geo-dist1}
Let $(\mathcal{M}, g)$ be a Riemannian manifold in an ambient $\mathbb{R}^n$, and let $p \in M$ be a point. If $g$ is locally constant about $p$ in an open neighbourhood $U$ such that $g$ is a constant diagonal matrix in ambient coordinates, then in a ball $B\subseteq U$ centered at $p$ with volume $\frac{\pi^{n/2}}{\Gamma(n/2 + 1)}$ with respect to $g$, the geodesic distance from $p$ to any point $q\in B$ is $\frac{1}{r} d_{\mathbb{R}^n}(p, q)$, where $r$ is the radius of the ball in the ambient space and $d_{\mathbb{R}^n}$ is the existing metric on the ambient space.
\end{lem}

See Appendix A of the supplementary materials for a proof of Lemma \ref{lem:geo-dist}.

If we assume the data to be uniformly distributed on $\mathcal{M}$ (with respect to $g$) then, away from any boundaries, any ball of fixed volume should contain approximately the same number of points of $X$ regardless of where on the manifold it is centered. Given finite data and small enough local neighborhoods this crude approximation should be accurate enough even for data samples near manifold boundaries. Now, conversely, a ball centered at $X_i$ that contains exactly the $k$-nearest-neighbors of $X_i$ should have approximately fixed volume regardless of the choice of $X_i \in X$. Under Lemma \ref{lem:geo-dist} it follows that we can approximate geodesic distance from $X_i$ to its neighbors by normalising distances with respect to the distance to the $k^{\text{th}}$ nearest neighbor of $X_i$.

In essence, by creating a custom distance for each $X_i$, we can ensure the validity of the assumption of uniform distribution on the manifold. The cost is that we now have an independent notion of distance for each and every $X_i$, and these notions of distance may not be compatible. We have a family of discrete metric spaces (one for each $X_i$) that we wish to merge into a consistent global structure. This can be done in a natural way by converting the metric spaces into fuzzy simplicial sets.

\subsection{Fuzzy topological representation}\label{kernel_derivation}

We will use functors between the relevant categories to convert from metric spaces to fuzzy topological representations. This will provide a means to merge the incompatible local views of the data. The topological structure of choice is that of simplicial sets. For more details on simplicial sets we refer the reader to \cite{goerss2009simplicial}, \cite{may1992simplicial}, \cite{riehl2011leisurely}, or \cite{friedman2012survey}. Our approach draws heavily upon the work of Michael Barr \cite{barr1986fuzzy} and David Spivak in \cite{spivakmetric}, and many of the definitions and theorems below are drawn or adapted from those sources. We assume familiarity with the basics of category theory. For an introduction to category theory readers may consult \cite{mac2013categories} or \cite{riehl2017category}.

To start we will review the definitions for simplicial sets. Simplicial sets provide a combinatorial approach to the study of topological spaces. They are related to the simpler notion of simplicial complexes -- which construct topological spaces by gluing together simple building blocks called simplices -- but are more general. Simplicial sets are most easily defined purely abstractly in the language of category theory.

\begin{defn}
The category $\boldsymbol{\Delta}$ has as objects the finite order sets $[n] = \{1, \ldots, n\}$, with morphims given by (non-strictly) order-preserving maps.
\end{defn}

Following standard category theoretic notation, $\boldsymbol{\Delta}^\text{op}$ denotes the category with the same objects as $\boldsymbol{\Delta}$ and morphisms given by the morphisms of $\boldsymbol{\Delta}$ with the direction (domain and codomain) reversed.

\begin{defn}
A \emph{simplicial set} is a functor from $\boldsymbol{\Delta}^\text{op}$ to \emph{\textbf{Sets}}, the category of sets; that is, a contravariant functor from $\boldsymbol{\Delta}$ to \emph{\textbf{Sets}}.
\end{defn}

Given a simplicial set $X:\boldsymbol{\Delta}^\text{op} \to \mathbf{Sets},$ it is common to denote the set $X([n])$ as $X_n$ and refer to the elements of the set as the $n$-simplices of $X$. The simplest possible examples of simplicial sets are the \emph{standard simplices} $\Delta^n$, defined as the representable functors $\hom_{\boldsymbol{\Delta}}(\cdot, [n])$. It follows from the Yoneda lemma that there is a natural correspondence between $n$-simplices of $X$ and morphisms $\Delta^n\to X$ in the category of simplicial sets, and it is often helpful to think in these terms.  Thus for each $x\in X_n$ we have a corresponding morphism $x:\Delta^n \to X$. By the density theorem and employing a minor abuse of notation we then have
\[
\colim_{x\in X_n} \Delta^n \cong X
\]

There is a standard covariant functor $|\cdot|: \boldsymbol{\Delta} \to \mathbf{Top}$ mapping from the category $\boldsymbol{\Delta}$ to the category of topological spaces that sends $[n]$ to the standard $n$-simplex $|\Delta^n| \subset \mathbb{R}^{n+1}$ defined as
\[
|\Delta^n| \triangleq \left\{(t_0, \ldots, t_n)\in \mathbb{R}^{n+1}\mid \sum_{i=0}^n t_i = 1, t_i \geq 0\right\}
\]
with the standard subspace topology.
If $X:\boldsymbol{\Delta}^\text{op} \to \mathbf{Sets}$ is a simplicial set then we can construct the realization of $X$ (denoted $|X|$) as the colimit
\[
|X| = \colim_{x\in X_n} |\Delta^n|
\]
and thus associate a topological space with a given simplicial set. Conversely given a topological space $Y$ we can construct an associated simplicial set $S(Y)$, called the singular set of $Y$, by defining 
\[
S(Y):[n]\mapsto \hom_{\mathbf{Top}}(|\Delta^n|,Y).
\]
It is a standard result of classical homotopy theory that the realization functor and singular set functors form an adjunction, and provide the standard means of translating between topological spaces and simplicial sets. Our goal will be to adapt these powerful classical results to the case of finite metric spaces.

We draw significant inspiration from Spivak, specifically \cite{spivakmetric}, where he extends the classical theory of singular sets and topological realization to fuzzy singular sets and metric realization. To develop this theory here we will first outline a categorical presentation of fuzzy sets, due to \cite{barr1986fuzzy}, that will make extending classical simplicial sets to fuzzy simplicial sets most natural. 

Classically a fuzzy set \cite{zadeh1965information} is defined in terms of a carrier set $A$ and a map $\mu:A\to [0,1]$ called the membership function. One is to interpret the value $\mu(x)$ for $x\in A$ to be the \emph{membership strength} of $x$ to the set $A$. Thus membership of a set is no longer a bi-valent true or false property as in classical set theory, but a fuzzy property taking values in the unit interval. We wish to formalize this in terms of category theory.

Let $I$ be the unit interval $(0,1] \subseteq \mathbb{R}$ with topology given by intervals of the form $[0,a)$ for $a\in (0, 1]$. The category of open sets (with morphisms given by inclusions) can be imbued with a Grothendieck topology in the natural way for any poset category.

\begin{defn}
A presheaf $\mathscr{P}$ on $I$ is a functor from $I^\text{op}$ to $\mathbf{Sets}$. A \emph{fuzzy set} is a presheaf on $I$ such that all maps $\mathscr{P}(a\leq b)$ are injections.
\end{defn}

Presheaves on $I$ form a category with morphisms given by natural transformations. We can thus form a category of fuzzy sets by simply restricting to the sub-category of presheaves that are fuzzy sets. We note that such presheaves are trivially sheaves under the Grothendieck topology on $I$. As one might expect, limits (including products) of such sheaves are well defined, but care must be taken to define colimits (and coproducts) of sheaves. To link to the classical approach to fuzzy sets one can think of a section $\mathscr{P}([0,a))$ as the set of all elements with membership strength at least $a$. We can now define the category of fuzzy sets.

\begin{defn}
The category $\mathbf{Fuzz}$ of \emph{fuzzy sets} is the full subcategory of sheaves on $I$ spanned by fuzzy sets.
\end{defn}

With this categorical presentation in hand, defining fuzzy simplicial sets is simply a matter of considering presheaves of $\boldsymbol\Delta$ valued in the category of fuzzy sets rather than the category of sets.

\begin{defn}
The category of \emph{fuzzy simplicial sets} $\mathbf{sFuzz}$ is the category with objects given by functors from $\boldsymbol\Delta^{\text{op}}$ to $\mathbf{Fuzz}$, and morphisms given by natural transformations.
\end{defn}

Alternatively, a fuzzy simplicial set can be viewed as a sheaf over $\boldsymbol\Delta \times I$, where $\boldsymbol\Delta$ is given the trivial topology and $\boldsymbol\Delta \times I$ has the product topology. We will use $\Delta^n_{<a}$ to denote the sheaf given by the representable functor of the object $([n], [0,a))$. The importance of this fuzzy (sheafified) version of simplicial sets is their relationship to metric spaces. We begin by considering the larger category of extended-pseudo-metric spaces.

\begin{defn}\label{defn:ex-pseu-met}
An \emph{extended-pseudo-metric space} $(X, d)$ is a set $X$ and a map $d: X\times X\to \mathbb{R}_{\geq 0}\cup\{\infty\}$ such that
\begin{enumerate}
    \item $d(x,y)\geqslant 0$, and $x=y$ implies $d(x,y)=0$;
    \item $d(x,y)=d(y,x)$; and
    \item $d(x,z)\leqslant d(x,y)+d(y,z)$ or $d(x,z)=\infty$.
\end{enumerate}
The category of extended-pseudo-metric spaces $\mathbf{EPMet}$ has as objects extended-pseudo-metric spaces and non-expansive maps as morphisms. We denote the subcategory of finite extended-pseudo-metric spaces $\mathbf{FinEPMet}$.
\end{defn}

The choice of non-expansive maps in Definition \ref{defn:ex-pseu-met} is due to Spivak, but we note that it closely mirrors the work of Carlsson and Memoli in \cite{carlsson2013classifying} on topological methods for clustering as applied to finite metric spaces. This choice is significant since pure isometries are too strict and do not provide large enough Hom-sets.

In \cite{spivakmetric} Spivak constructs a pair of adjoint functors, $\Real$ and $\Sing$ between the categories \textbf{sFuzz} and \textbf{EPMet}. These functors are the natural extension of the classical realization and singular set functors from algebraic topology. The functor $\Real$ is defined in terms of standard fuzzy simplices $\Delta^n_{<a}$ as
\[
\Real(\Delta^n_{<a}) \triangleq \left\{ (t_0, \ldots, t_n)\in \mathbb{R}^{n+1}\mid \sum_{i=0}^n t_i = -\log(a), t_i \geq 0 \right\}
\]
similarly to the classical realization functor $|\cdot|$. The metric on $\Real(\Delta^n_{<a})$ is simply inherited from $\mathbb{R}^{n+1}$. A morphism $\Delta^n_{<a}\to \Delta^m_{<b}$ exists only if $a\leq b$, and is determined by a $\boldsymbol{\Delta}$ morphism  $\sigma:[n]\to [m]$. The action of $\Real$ on such a morphism is given by the map
\[
(x_0,x_1,\ldots,x_n) \mapsto \frac{\log(b)}{\log(a)}\left(\sum_{i_0\in\sigma^{-1}(0)} x_{i_0}, \sum_{i_0\in\sigma^{-1}(1)} x_{i_0}, \ldots, \sum_{i_0\in\sigma^{-1}(m)} x_{i_0}\right).
\]
Such a map is clearly non-expansive since $0\leq a \leq b \leq 1$ implies that $\log(b)/\log(a) \leq 1$.

We then extend this to a general simplicial set $X$ via colimits, defining
\[
\Real(X) \triangleq \colim_{\Delta^n_{<a}\to X} \Real(\Delta^n_{<a}).
\]

Since the functor $\Real$ preserves colimits, it follows that there exists a right adjoint functor. Again, analogously to the classical case, we find the right adjoint, denoted $\Sing$, is defined for an extended pseudo metric space $Y$ in terms of its action on the category $\boldsymbol\Delta \times I$:
\[
\Sing(Y):([n], [0,a)) \mapsto \hom_{\textbf{EPMet}}(\Real(\Delta^n_{<a}), Y).
\]

For our case we are only interested in finite metric spaces. To correspond with this we consider the subcategory of bounded fuzzy simplicial sets \textbf{Fin-sFuzz}. We therefore use the analogous adjoint pair $\FinReal$ and $\FinSing$. Formally we define the finite fuzzy realization functor as follows:

\begin{defn}\label{defn:fin-real-func}
Define the functor $\FinReal:\text{\rm \bf Fin-sFuzz}\to\mathbf{FinEPMet}$ by setting
\[
\FinReal(\Delta^n_{<a}) \triangleq (\{x_1, x_2, \ldots, x_n\}, d_a),
\]
where
\[
d_a(x_i, x_j) = \begin{cases}
    -\log(a) & \quad\text{if } i \neq j,\\[4pt]
    0 & \quad\text{otherwise}
\end{cases}.
\]
and then defining
\[
\FinReal(X) \triangleq \colim_{\Delta^n_{<a}\to X} \FinReal(\Delta^n_{<a}).
\]
\end{defn}
Similar to Spivak's construction, the action of $\FinReal$ on a map $\Delta^n_{<a}\to \Delta^m_{<b}$, where $a\leq b$ defined by $\sigma:\Delta^n\to\Delta^m$, is given by
\[
(\{x_1, x_2, \ldots, x_n\}, d_a)\mapsto (\{x_{\sigma(1)}, x_{\sigma(2)}, \ldots, x_{\sigma(n)}\}, d_b),
\] 
which is a non-expansive map since $a \leq b$ implies $d_a \geq d_b$.

Since $\FinReal$ preserves colimits it admits a right adjoint, the fuzzy singular set functor $\FinSing$. We can then define the (finite) fuzzy singular set functor in terms of the action of its image on $\boldsymbol\Delta \times I$, analogously to $\Sing$.

\begin{defn}
Define the functor $\FinSing:\mathbf{FinEPMet}\to\text{\rm \bf Fin-sFuzz}$ by
\[
\FinSing(Y): ([n], [0,a)) \mapsto \hom_{\text{\bf FinEPMet}}(\FinReal(\Delta^n_{<a}), Y).
\]
\end{defn}

We then have the following theorem.
\begin{thm}\label{thm:adjunction}
The functors $\FinReal:\text{\rm \bf Fin-sFuzz}\to\mathbf{FinEPMet}$ and $\FinSing:\mathbf{FinEPMet}\to \text{\rm \bf Fin-sFuzz}$ form an adjunction with $\FinReal$ the left adjoint and $\FinSing$ the right adjoint.
\end{thm}

The proof of this is by construction. Appendix B provides a full proof of the theorem.

With the necessary theoretical background in place, the means to handle the family of incompatible metric spaces described above becomes clear. Each metric space in the family can be translated into a fuzzy simplicial set via the fuzzy singular set functor, distilling the topological information while still retaining metric information in the fuzzy structure. Ironing out the incompatibilities of the resulting family of fuzzy simplicial sets can be done by simply taking a (fuzzy) union across the entire family. The result is a single fuzzy simplicial set which captures the relevant topological and underlying metric structure of the manifold $\mathcal{M}$.

It should be noted, however, that the fuzzy singular set functor applies to extended-pseudo-metric spaces, which are a relaxation of traditional metric spaces. The results of Lemma \ref{lem:geo-dist} only provide accurate approximations of geodesic distance local to $X_i$ for distances measured from $X_i$ -- the geodesic distances between other pairs of points within the neighborhood of $X_i$ are not well defined. In deference to this lack of information we define distances between $X_j$ and $X_k$ in the extended-pseudo metric space local to $X_i$ (where $i\neq j$ and $i\neq k$) to be infinite (local neighborhoods of $X_j$ and $X_k$ will provide suitable approximations).

For real data it is safe to assume that the manifold $\mathcal{M}$ is locally connected. In practice this can be realized by measuring distance in the extended-pseudo-metric space local to $X_i$ as geodesic distance \emph{beyond} the nearest neighbor of $X_i$. Since this sets the distance to the nearest neighbor to be equal to 0 this is only possible in the more relaxed setting of extended-pseudo-metric spaces. It ensures, however, that each 0-simplex is the face of some 1-simplex with fuzzy membership strength 1, meaning that the resulting topological structure derived from the manifold is locally connected. We note that this has a similar practical effect to the truncated similarity approach of Lee and Verleysen \cite{lee2011shift}, but derives naturally from the assumption of local connectivity of the manifold.

Combining all of the above we can define the fuzzy topological representation of a dataset.

\begin{defn}\label{defn:fuzz-topo-repr}
Let $X = \{X_1, \ldots, X_N\}$ be a dataset in $\mathbb{R}^n$. Let $\{(X, d_i)\}_{i=1\ldots N}$ be a family of extended-pseudo-metric spaces with common carrier set $X$ such that
\[
d_i(X_j, X_k) = \begin{cases}
       d_{\mathcal{M}}(X_j, X_k)  - \rho &  \text{ if $i = j$ or $i = k$},\\[8pt]
        \infty & \text{ otherwise },
    \end{cases}
\]
where $\rho$ is the distance to the nearest neighbor of $X_i$ and $d_{\mathcal{M}}$ is geodesic distance on the manifold $\mathcal{M}$, either known apriori, or approximated as per Lemma \ref{lem:geo-dist}.

The fuzzy topological representation of $X$ is
\[
\bigcup_{i=1}^n \FinSing((X, d_i)).
\]
\end{defn}

The (fuzzy set) union provides the means to merge together the different metric spaces. This provides a single fuzzy simplicial set as the global representation of the manifold formed by patching together the many local representations.

Given the ability to construct such topological structures, either from a known manifold, or by learning the metric structure of the manifold, we can perform dimension reduction by simply finding low dimensional representations that closely match the topological structure of the source data. We now consider the task of finding such a low dimensional representation.

\subsection{Optimizing a low dimensional representation}\label{graph_layout_derivation}

Let $Y = \{Y_1,\ldots,Y_N\} \subseteq \mathbb{R}^d$ be a low dimensional ($d \ll n$) representation of $X$ such that $Y_i$ represents the source data point $X_i$. In contrast to the source data where we want to estimate a manifold on which the data is uniformly distributed, a target manifold for $Y$ is chosen apriori (usually this will simply be $\mathbb{R}^d$ itself, but other choices such as $d$-spheres or $d$-tori are certainly possible) . Therefore we know the manifold and manifold metric apriori, and can compute the fuzzy topological representation directly. Of note, we still want to incorporate the distance to the nearest neighbor as per the local connectedness requirement. This can be achieved by supplying a parameter that defines the expected distance between nearest neighbors in the embedded space.

Given fuzzy simplicial set representations of $X$ and $Y$, a means of comparison is required. If we consider only the 1-skeleton of the fuzzy simplicial sets we can describe each as a fuzzy graph, or, more specifically, a fuzzy set of edges. To compare two fuzzy sets we will make use of fuzzy set cross entropy. For these purposes we will revert to classical fuzzy set notation. That is, a fuzzy set is given by a reference set $A$ and a membership strength function $\mu:A\to[0,1]$. Comparable fuzzy sets have the same reference set. Given a sheaf representation $\mathscr{P}$ we can translate to classical fuzzy sets by setting $A = \bigcup_{a\in(0,1]} \mathscr{P}([0,a))$ and $\mu(x) = \sup \{a\in (0, 1] \mid x \in \mathscr{P}([0, a))\}$. 

\begin{defn}
The cross entropy $C$ of two fuzzy sets $(A, \mu)$ and $(A, \nu)$ is defined as
\[
C((A, \mu), (A, \nu)) \triangleq \sum_{a\in A} \left(\mu(a)\log\left(\frac{\mu(a)}{\nu(a)}\right)
     + (1 - \mu(a))\log\left(\frac{1 - \mu(a)}{1 - \nu(a)}\right)\right) .
\]
\end{defn}

Similar to t-SNE we can optimize the embedding $Y$ with respect to fuzzy set cross entropy $C$ by using stochastic gradient descent. However, this requires a differentiable fuzzy singular set functor. If the expected minimum distance between points is zero the fuzzy singular set functor is differentiable for these purposes, however for any non-zero value we need to make a differentiable approximation (chosen from a suitable family of differentiable functions).

This completes the algorithm: by using manifold approximation and patching together local fuzzy simplicial set representations we construct a topological representation of the high dimensional data. We then optimize the layout of data in a low dimensional space to minimize the error between the two topological representations.

We note that in this case we restricted attention to comparisons of the 1-skeleton of the fuzzy simplicial sets. One can extend this to $\ell$-skeleta by defining a cost function $C_\ell$ as 
\[
C_\ell(X, Y) = \sum_{i=1}^\ell \lambda_i C(X_i, Y_i),
\]
where $X_i$ denotes the fuzzy set of $i$-simplices of $X$ and the $\lambda_i$ are suitably chosen real valued weights. While such an approach will capture the overall topological structure more accurately, it comes at non-negligible computational cost due to the increasingly large numbers of higher dimensional simplices. For this reason current implementations restrict to the 1-skeleton at this time.

\section{A Computational View of UMAP}\label{computational}

To understand what computations the UMAP algorithm is actually making from a practical point of view, a less theoretical and more computational description may be helpful for the reader.  This description of the algorithm lacks the motivation for a number of the choices made.  For that motivation please see Section \ref{theory}.

The theoretical description of the algorithm works in terms of fuzzy simplicial sets.  Computationally this is only tractable for the one skeleton which can ultimately be described as a weighted graph.  This means that, from a practical computational perspective, UMAP can ultimately be described in terms of, construction of, and operations on, weighted graphs.  In particular this situates UMAP in the class of k-neighbour based graph learning algorithms such as Laplacian Eigenmaps, Isomap and t-SNE.

As with other k-neighbour graph based algorithms, UMAP can be described in two phases.  In the first phase a particular weighted k-neighbour graph is constructed.  In the second phase a low dimensional layout of this graph is computed.  The differences between all algorithms in this class amount to specific details in how the graph is constructed and how the layout is computed.  The theoretical basis for UMAP as described in Section \ref{theory} provides novel approaches to both of these phases, and provides clear motivation for the choices involved.  

Finally, since t-SNE is not usually described as a graph based algorithm, a direct comparison of UMAP with t-SNE, using the similarity/probability notation commonly used to express the equations of t-SNE, is given in the Appendix \ref{compare}.

In section {\ref{theory}} we made a few basic assumptions about our data. From these assumptions we made use of category theory to derive the UMAP algorithms. That said, all these derivations assume these axioms to be true. 
\begin{enumerate}
    \item There exists a manifold on which the data would be uniformly distributed.
    \item The underlying manifold of interest is locally connected.
    \item Preserving the topological structure of this manifold is the primary goal.
\end{enumerate}
The topological theory of Section {\ref{theory}} is driven by these axioms, particularly the interest in modelling and preserving topological structure. In particular Section {\ref{graph_derivation}} highlights the underlying motivation, in terms of topological theory, of representing a manifold as a k-neighbour graph.

As highlighted in Appendix {\ref{compare}} any algorithm that attempts to use a mathematical structure akin to a k-neighbour graph to approximate a manifold must follow a similar basic structure.

\begin{itemize}
    \item Graph Construction
    \vspace*{-0.8em}
    \begin{enumerate}
        \item Construct a weighted k-neighbour graph
        \item Apply some transform on the edges to ambient local distance.
        \item Deal with the inherent asymmetry of the k-neighbour graph.
    \end{enumerate}
    \item Graph Layout
    \vspace*{-0.8em}
    \begin{enumerate}
        \item Define an objective function that preserves desired characteristics of this k-neighbour graph.
        \item Find a low dimensional representation which optimizes this objective function.       
    \end{enumerate}
\end{itemize}

Many dimension reduction algorithms can be broken down into these steps because they are fundamental to a particular class of solutions.  Choices for each step must be either chosen through task oriented experimentation or by selecting a set of believable axioms and building strong theoretical arguments from these.  Our belief is that basing our decisions on a strong foundational theory will allow for a more extensible and generalizable algorithm in the long run.

We theoretically justify using the choice of using a k-neighbour graph to represent a manifold in Section {\ref{graph_derivation}}.  The choices for our kernel transform an symmetrization function can be found in Section {\ref{kernel_derivation}}. Finally, the justifications underlying our choices for our graph layout are outlined in Section {\ref{graph_layout_derivation}}.

\subsection{Graph Construction}

The first phase of UMAP can be thought of as the construction of a weighted k-neighbour graph. Let $X  = \{x_1,\ldots,x_N\}$ be the input dataset, with a metric (or dissimilarity measure) $d:X\times X \to \mathbb{R}_{\geq 0}$. Given an input hyper-parameter $k$, for each $x_i$ we compute the set $\{x_{i_1},\ldots, x_{i_k}\}$ of the $k$ nearest neighbors of $x_i$ under the metric $d$. This computation can be performed via any nearest neighbour or approximate nearest neighbour search algorithm.  For the purposes of our UMAP implemenation we prefer to use the nearest neighbor descent algorithm of \cite{DongNearestNeighbour}.

For each $x_i$ we will define $\rho_i$ and $\sigma_i$. Let 
\[
\rho_i = \min \{ d(x_i, x_{i_j}) \mid 1\leq j\leq k, d(x_i, x_{i_j}) > 0\},
\]
and set $\sigma_i$ to be the value such that
\[
\sum_{j=1}^k \exp\left(\frac{-\max(0, d(x_i, x_{i_j}) - \rho_i)}{\sigma_i}\right) = \log_2(k).
\]
The selection of $\rho_i$ derives from the local-connectivity constraint described in Section {\ref{kernel_derivation}}. In particular it ensures that $x_i$ connects to at least one other data point with an edge of weight 1; this is equivalent to the resulting fuzzy simplicial set being locally connected at $x_i$. In practical terms this significantly improves the representation on very high dimensional data where other algorithms such as t-SNE begin to suffer from the curse of dimensionality.

The selection of $\sigma_i$ corresponds to (a smoothed) normalisation factor, defining the Riemannian metric local to the point $x_i$ as described in Section {\ref{graph_derivation}}.

We can now define a weighted directed graph $\bar{G} = (V, E, w)$. The vertices $V$ of $\bar{G}$ are simply the set $X$. We can then form the set of directed edges $E = \{(x_i, x_{i_j}) \mid 1\leq j\leq k, 1\leq i\leq N\}$, and define the weight function $w$ by setting
\[
w((x_i, x_{i_j})) = \exp\left(\frac{-\max(0, d(x_i, x_{i_j}) - \rho_i)}{\sigma_i}\right).
\]
For a given point $x_i$ there exists an induced graph of $x_i$ and outgoing edges incident on $x_i$. This graph is the 1-skeleton of the fuzzy simplicial set associated to the metric space local to $x_i$ where the local metric is defined in terms of $\rho_i$ and $\sigma_i$. The weight associated to the edge is the membership strength of the corresponding 1-simplex within the fuzzy simplicial set, and is derived from the adjunction of Theorem {\ref{thm:adjunction}} using the right adjoint (nearest inverse) of the geometric realization of a fuzzy simplicial set. Intuitively one can think of the weight of an edge as akin to the probability that the given edge exists. Section \ref{theory} demonstrates why this construction faithfully  captures the topology of the data. Given this set of local graphs (represented here as a single directed graph) we now require a method to combine them into a unified topological representation. We note that while patching together incompatible finite metric spaces is challenging, by using Theorem {\ref{thm:adjunction}} to convert to a fuzzy simplicial set representation, the combining operation becomes natural.

Let $A$ be the weighted adjacency matrix of $\bar{G}$, and consider the symmetric matrix
\[
B = A + A^\top - A \circ A^\top,
\]
where $\circ$ is the Hadamard (or pointwise) product. This formula derives from the use of the probabilistic t-conorm used in unioning the fuzzy simplicial sets. If one interprets the value of $A_{ij}$ as the probability that the directed edge from $x_i$ to $x_j$ exists, then $B_{ij}$ is the probability that at least one of the two directed edges (from $x_i$ to $x_j$ and from $x_j$ to $x_i$) exists. The UMAP graph $G$ is then an undirected weighted graph whose adjacency matrix is given by $B$. Section \ref{theory} explains this construction in topological terms, providing the justification for why this construction provides an appropriate fuzzy topological representation of the data -- that is, this construction captures the underlying geometric structure of the data in a faithful way.

\subsection{Graph Layout}

In practice UMAP uses a force directed graph layout algorithm in low dimensional space.  A force directed graph layout utilizes of a set of attractive forces applied along edges and a set of repulsive forces applied among vertices.  Any force directed layout algorithm requires a description of both the attractive and repulsive forces.  The algorithm proceeds by iteratively applying attractive and repulsive forces at each edge or vertex.  This amounts to a non-convex optimization problem. Convergence to a local minima is guaranteed by slowly decreasing the attractive and repulsive forces in a similar fashion to that used in simulated annealing.  

In UMAP the attractive force between two vertices $i$ and
$j$ at coordinates $\mathbf{y_i}$ and $\mathbf{y_j}$ respectively, is determined by:

\[
\frac{-2ab \|\mathbf{y_i} - \mathbf{y_j} \|_2^
{2\left(b - 1\right)}}{1 + \|\mathbf{y_i} - \mathbf{y_j} \|_2^2}  w((x_i, x_j)) \left(\mathbf{y_i - y_j}\right)
\]\label{Attractive equation}

where $a$ and $b$ are hyper-parameters.

Repulsive forces are computed via sampling due to computational constraints.  Thus, whenever an attractive force is applied to an edge, one of that edge's vertices is repulsed by a sampling of other vertices.  The repulsive force is given by

\[
\frac{2b}{\left(\epsilon + \|\mathbf{y_i} - \mathbf{y_j} \|_2^2\right)
\left(1 + a\|\mathbf{y_i} - \mathbf{y_j} \|_2^{2b}\right)}
\left(1 - w((x_i, x_j))\right)\left(\mathbf{y_i - y_j}\right).
\]\label{Repulsive equation}

$\epsilon$ is a small number to prevent division by zero (0.001 in the current implementation).

The algorithm can be initialized randomly but in practice, since the symmetric Laplacian of the graph $G$ is a discrete approximation of the Laplace-Beltrami operator of the manifold, we can use a spectral layout to initialize the embedding.  This provides both faster convergence and greater stability within the algorithm.

The forces described above are derived from gradients optimising the edge-wise cross-entropy between the weighted graph $G$, and an equivalent weighted graph $H$ constructed from the points $\{\mathbf{y_i}\}_{i=1..N}$. That is, we are seeking to position points $y_i$ such that the weighted graph induced by those points most closely approximates the graph $G$, where we measure the difference between weighted graphs by the total cross entropy over all the edge existence probabilities. Since the weighted graph $G$ captures the topology of the source data, the equivalent weighted graph $H$ constructed from the points $\{\mathbf{y_i}\}_{i=1..N}$ matches the topology as closely as the optimization allows, and thus provides a good low dimensional representation of the overall topology of the data.

\section{Implementation and Hyper-parameters}\label{implementation}

Having completed a theoretical description of the approach, we now turn our attention to the practical realization of this theory. We begin by providing a more detailed description of the algorithm as implemented, and then discuss a few implementation specific details. We conclude this section with a discussion of the hyper-parameters for the algorithm and their practical effects.

\subsection{Algorithm description}\label{algorithm}

In overview the UMAP algorithm is relatively straightforward (see Algorithm \ref{alg:umap}). When performing a fuzzy union over local fuzzy simplicial sets we have found it most effective to work with the probabilistic t-conorm (as one would expect if treating membership strengths as a probability that the simplex exists). The individual functions for constructing the local fuzzy simplicial sets, determining the spectral embedding, and optimizing the embedding with regard to fuzzy set cross entropy, are described in more detail below.

\begin{algorithm}[!htbp]
\caption{UMAP algorithm}\label{alg:umap}
\begin{algorithmic}[0]
\setlength\baselineskip{18pt}
\Function{UMAP}{$X$, $n$, $d$, min-dist, n-epochs}
    \State
    \State \# \textit{Construct the relevant weighted graph}
    \ForAll{$x \in X$}
        \State fs-set[$x$] $\gets$ \Call{LocalFuzzySimplicialSet}{$X$, $x$, $n$}
    \EndFor
    \State top-rep $\gets \bigcup_{x\in X} \textrm{fs-set}[x]$ \Comment{We recommend the probabilistic t-conorm}
    \State
    \State \# \textit{Perform optimization of the graph layout}
    \State $Y \gets$ \Call{SpectralEmbedding}{top-rep, $d$}
    \State $Y \gets$ \Call{OptimizeEmbedding}{top-rep, $Y$, min-dist, n-epochs}
    \State \Return $Y$
\EndFunction\vskip9pt
\end{algorithmic}
\end{algorithm}

The inputs to Algorithm \ref{alg:umap} are: $X$, the dataset to have its dimension reduced; $n$, the neighborhood size to use for local metric approximation; $d$, the dimension of the target reduced space; min-dist, an algorithmic parameter controlling the layout; and n-epochs, controlling the amount of optimization work to perform.

Algorithm \ref{alg:fss-construction} describes the construction of local fuzzy simplicial sets. To represent fuzzy simplicial sets we work with the fuzzy set images of $[0]$ and $[1]$ (i.e. the 1-skeleton), which we denote as $\text{fs-set}_0$ and $\text{fs-set}_1$. One can work with higher order simplices as well, but the current implementation does not. We can construct the fuzzy simplicial set local to a given point $x$ by finding the $n$ nearest neighbors, generating the appropriate normalised distance on the manifold, and then converting the finite metric space to a simplicial set via the functor $\FinSing$, which translates into exponential of the negative distance in this case.

\begin{algorithm}[!htbp]
\caption{Constructing a local fuzzy simplicial set}\label{alg:fss-construction}
\begin{algorithmic}[0]
\setlength\baselineskip{18pt}
\Function{LocalFuzzySimplicialSet}{$X$, $x$, $n$}
    \State knn, knn-dists $\gets$ \Call{ApproxNearestNeighbors}{$X$, $x$, $n$}
    \State $\rho \gets $ knn-dists[1] \Comment{Distance to nearest neighbor}
    \State $\sigma \gets $ \Call{SmoothKNNDist}{knn-dists, $n$, $\rho$}
    \Comment{Smooth approximator to knn-distance}
    \State $\text{fs-set}_0 \gets X$
    \State $\text{fs-set}_1 \gets \{([x,y], 0) \mid y \in X\}$
    \ForAll {$y \in$ knn}
        \State $d_{x,y} \gets \max\{0, \textrm{dist}(x, y) - \rho\}/\sigma$
        \State $\text{fs-set}_1 \gets \text{fs-set}_1 \cup\, ([x, y], \exp(-d_{x,y}))$
    \EndFor
    \State \Return fs-set
\EndFunction\vskip9pt
\end{algorithmic}
\end{algorithm}

Rather than directly using the distance to the $n^\text{th}$ nearest neighbor as the normalization, we use a smoothed version of knn-distance that fixes the cardinality of the fuzzy set of 1-simplices to a fixed value. We selected $\log_2(n)$ for this purpose based on empirical experiments. This is described briefly in Algorithm \ref{alg:smooth-knn}.

\begin{algorithm}[!htbp]
\caption{Compute the normalizing factor for distances $\sigma$}\label{alg:smooth-knn}
\begin{algorithmic}[0]
\setlength\baselineskip{18pt}
\Function{SmoothKNNDist}{knn-dists, $n$, $\rho$}
    \State Binary search for $\sigma$ such that $\sum_{i=1}^n \exp(-(\text{knn-dists}_i - \rho)/\sigma) = \log_2(n)$
    \State \Return $\sigma$
\EndFunction\vskip9pt
\end{algorithmic}
\end{algorithm}

Spectral embedding is performed by considering the 1-skeleton of the global fuzzy topological representation as a weighted graph and using standard spectral methods on the symmetric normalized Laplacian. This process is described in Algorithm \ref{alg:init-embed}.

\begin{algorithm}[!htbp]
\caption{Spectral embedding for initialization}\label{alg:init-embed}
\begin{algorithmic}[0]
\setlength\baselineskip{18pt}
\Function{SpectralEmbedding}{top-rep, $d$}
    \State $A \gets$ 1-skeleton of top-rep expressed as a weighted adjacency matrix
    \State $D \gets$ degree matrix for the graph $A$
    \State $L \gets D^{1/2}(D - A)D^{1/2}$
    \State $\text{evec} \gets$ Eigenvectors of $L$ (sorted)
    \State $Y \gets \text{evec}[1..d+1]$ \Comment{0-base indexing assumed}
    \State \Return $Y$
\EndFunction\vskip9pt
\end{algorithmic}
\end{algorithm}

The final major component of UMAP is the optimization of the embedding through minimization of the fuzzy set cross entropy. Recall that fuzzy set cross entropy, with respect given membership functions $\mu$ and $\nu$, is given by
\begin{equation}
    \begin{split}
        C((A, \mu), (A, \nu)) = & \sum_{a\in A} \mu(a)\log\left(\frac{\mu(a)}{\nu(a)}\right)
                      + (1 - \mu(a))\log\left(\frac{1 - \mu(a)}{1 - \nu(a)}\right)\\
        = & \sum_{a\in A} \left(\mu(a)\log(\mu(a)) + (1 - \mu(a))\log(1 - \mu(a))\right)\\
          & \qquad - \sum_{a\in A} \left(\mu(a)\log(\nu(a)) + (1 - \mu(a))\log(1 - \nu(a))\right).
    \end{split}
\end{equation}
The first sum depends only on $\mu$ which takes fixed values during the optimization, thus the minimization of cross entropy depends only on the second sum, so we seek to minimize
\[
- \sum_{a\in A} \left(\mu(a)\log(\nu(a)) + (1 - \mu(a))\log(1 - \nu(a))\right).
\]
Following both \cite{tang2016visualizing} and \cite{mikolov2013distributed}, we take a sampling based approach to the optimization. We sample 1-simplices with probability $\mu(a)$ and update according to the value of $\nu(a)$, which handles the term $\mu(a)\log(\nu(a))$. The term $(1 - \mu(a))\log(1 - \nu(a))$ requires negative sampling -- rather than computing this over all potential simplices we randomly sample potential 1-simplices and assume them to be a negative example (i.e. with membership strength 0) and update according to the value of $1 - \nu(a)$. In contrast to \cite{tang2016visualizing} the above formulation provides a vertex sampling distribution of
\[
P(x_i) = \frac{\sum_{\{a\in A\mid d_0(a) = x_i \}} 1 - \mu(a)}{\sum_{\{b\in A\mid d_0(b) \neq x_i\}} 1 - \mu(b)}
\]
for negative samples, which can be reasonably approximated by a uniform distribution for sufficiently large data sets.

It therefore only remains to find a differentiable approximation to $\nu(a)$ for a given 1-simplex $a$ so that gradient descent can be applied for optimization. This is done as follows:

\begin{defn}\label{defn:nu_approx}
Define $\Phi:\mathbb{R}^d \times \mathbb{R}^d \to [0, 1]$, a smooth approximation of the membership strength of a 1-simplex between two points in $\mathbb{R}^d$, as
\[
\Phi(\mathbf{x},\mathbf{y}) = \left(1 + a (\|\mathbf{x} - \mathbf{y} \|_2^2)^{b}\right)^{-1},
\]
where $a$ and $b$ are chosen by non-linear least squares fitting against the curve $\Psi:\mathbb{R}^d \times \mathbb{R}^d \to [0, 1]$ where
\[
\Psi(\mathbf{x},\mathbf{y}) = \begin{cases}
1 & \text{if } \|\mathbf{x} - \mathbf{y}\|_2 \leq \text{min-dist}\\
\exp(-(\|\mathbf{x} - \mathbf{y}\|_2 - \text{min-dist})) & \text{otherwise}
\end{cases}.
\]
\end{defn}

The optimization process is now executed by stochastic gradient descent as given by Algorithm \ref{alg:opt-embed}.

\begin{algorithm}[!hbpt]
\caption{Optimizing the embedding}\label{alg:opt-embed}
\begin{algorithmic}[0]
\setlength\baselineskip{18pt}
\Function{OptimizeEmbedding}{top-rep, $Y$, min-dist, n-epochs}
    \State $\alpha \gets 1.0$
    \State Fit $\Phi$ from $\Psi$ defined by min-dist
    \For{$e \gets 1,\ldots,$ n-epochs}
        \ForAll{$([a,b], p) \in \text{top-rep}_1$}
            \If{\Call{Random}{ }~$ \leq p$} \Comment{Sample simplex with probability $p$}
                \State $y_a  \gets y_a + \alpha \cdot \nabla (\log(\Phi)) (y_a, y_b)$
                \For {$i \gets 1,\ldots,\text{n-neg-samples}$}
                    \State $c \gets \text{random sample from Y}$
                    \State $y_a \gets y_a + \alpha \cdot \nabla (\log(1 - \Phi)) (y_a, y_c)$
                \EndFor
            \EndIf
        \EndFor
        \State
        \State $\alpha \gets 1.0 - e/\text{n-epochs}$
    \EndFor
    \State \Return $Y$
\EndFunction\vskip9pt
\end{algorithmic}
\end{algorithm}

This completes the UMAP algorithm.

\subsection{Implementation}\label{implementation-detail}

Practical implementation of this algorithm requires (approximate) $k$-nearest-neighbor calculation and efficient optimization via stochastic gradient descent. 

Efficient approximate $k$-nearest-neighbor computation can be achieved via the Nearest-Neighbor-Descent algorithm of \cite{DongNearestNeighbour}. The error intrinsic in a dimension reduction technique means that such approximation is more than adequate for these purposes. While no theoretical complexity bounds have been established for Nearest-Neighbor-Descent the authors of the original paper report an empirical complexity of $O(N^{1.14})$. A further benefit of Nearest-Neighbor-Descent is its generality; it works with any valid dissimilarity measure, and is efficient even for high dimensional data.

In optimizing the embedding under the provided objective function, we follow work of \cite{tang2016visualizing}; making use of probabilistic edge sampling and negative sampling \cite{mikolov2013distributed}. This provides a very efficient approximate stochastic gradient descent algorithm since there is no normalization requirement. Furthermore, since the normalized Laplacian of the fuzzy graph representation of the input data is a discrete approximation of the Laplace-Betrami operator of the manifold \cite[see][]{belkin2002laplacian, belkin2003laplacian}, we can provide a suitable initialization for stochastic gradient descent by using the eigenvectors of the normalized Laplacian. The amount of optimization work required will scale with the number of edges in the fuzzy graph (assuming a fixed negative sampling rate), resulting in a complexity of $O(kN)$.

Combining these techniques results in highly efficient embeddings, which we will discuss in Section \ref{experiments}. The overall complexity is bounded by the approximate nearest neighbor search complexity and, as mentioned above, is empirically approximately $O(N^{1.14})$. A reference implementation can be found at \url{https://github.com/lmcinnes/umap}, and an R implementation can be found at \url{https://github.com/jlmelville/uwot}.

For simplicity these experiments were carried out on a single core version of our algorithm.  It should be noted that at the time of this publication that both Nearest-Neighbour-Descent and SGD have been parallelized and thus the python reference implementation can be significantly accelerated.   Our intention in this paper was to introduce the underlying theory behind our UMAP algorithm and we felt that parallel vs single core discussions would distract from our intent.

\subsection{Hyper-parameters}

As described in Algorithm \ref{alg:umap}, the UMAP algorithm takes four hyper-parameters:
\begin{enumerate}
    \item $n$, the number of neighbors to consider when approximating the local metric;
    \item $d$, the target embedding dimension;
    \item min-dist, the desired separation between close points in the embedding space; and
    \item n-epochs, the number of training epochs to use when optimizing the low dimensional representation.
\end{enumerate}   
The effects of the parameters $d$ and n-epochs are largely self-evident, and will not be discussed in further detail here. In contrast the effects of the number of neighbors $n$ and of min-dist are less clear.

One can interpret the number of neighbors $n$ as the local scale at which to approximate the manifold as roughly flat, with the manifold estimation averaging over the $n$ neighbors. Manifold features that occur at a smaller scale than within the $n$ nearest-neighbors of points will be lost, while large scale manifold features that cannot be seen by patching together locally flat charts at the scale of $n$ nearest-neighbors may not be well detected. Thus $n$ represents some degree of trade-off between fine grained and large scale manifold features --- smaller values will ensure detailed manifold structure is accurately captured (at a loss of the ``big picture'' view of the manifold), while larger values will capture large scale manifold structures, but at a loss of fine detail structure which will get averaged out in the local approximations. With smaller $n$ values the manifold tends to be broken into many small connected components (care needs to be taken with the spectral embedding for initialization in such cases).

In contrast min-dist is a hyperparameter directly affecting the output, as it controls the fuzzy simplicial set construction from the low dimensional representation. It acts in lieu of the distance to the nearest neighbor used to ensure local connectivity. In essence this determines how closely points can be packed together in the low dimensional representation. Low values on min-dist will result in potentially densely packed regions, but will likely more faithfully represent the manifold structure. Increasing the value of min-dist will force the embedding to spread points out more, assisting visualization (and avoiding potential overplotting issues). We view min-dist as an essentially aesthetic parameter, governing the appearance of the embedding, and thus is more important when using UMAP for visualization.

\begin{figure}
    \centering
    \includegraphics[width=0.9\textwidth]{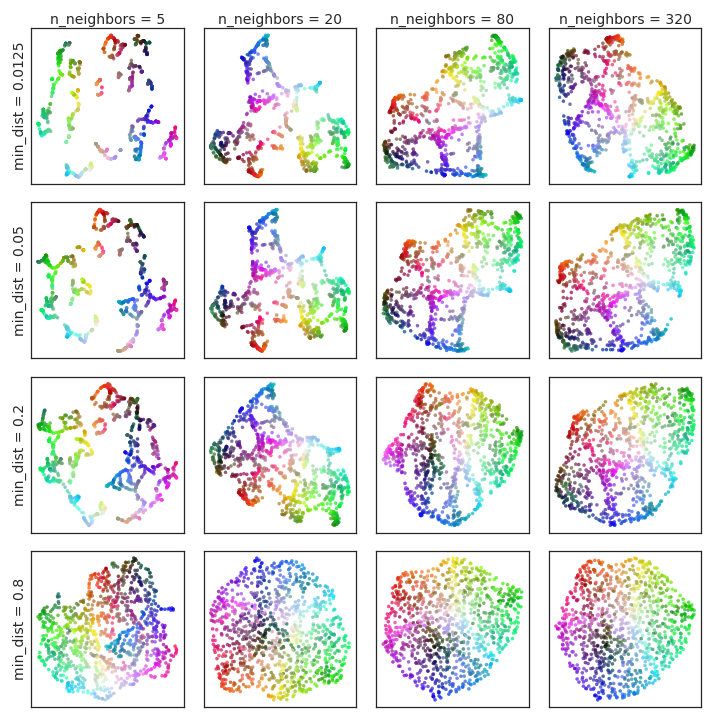}
    \caption{Variation of UMAP hyperparameters $n$ and min-dist result in different embeddings. The data is uniform random samples from a 3-dimensional color-cube, allowing for easy visualization of the original 3-dimensional coordinates in the embedding space by using the corresponding RGB colour. Low values of $n$ spuriously interpret structure from the random sampling noise -- see Section \ref{weaknesses} for further discussion of this phenomena.}
    \label{fig:hyperparameters}
\end{figure}

In Figure \ref{fig:hyperparameters} we provide examples of the effects of varying the hyperparameters for a toy dataset. The data is uniform random samples from a 3-dimensional color-cube, allowing for easy visualization of the original 3-dimensional coordinates in the embedding space by using the corresponding RGB colour. Since the data fills a 3-dimensional cube there is no local manifold structure, and hence for such data we expect larger $n$ values to be more useful. Low values will interpret the noise from random sampling as fine scale manifold structure, producing potentially spurious structure\footnote{See the discussion of the constellation effect in Section \ref{weaknesses}}.

\begin{figure}
    \centering
    \includegraphics[width=0.9\textwidth]{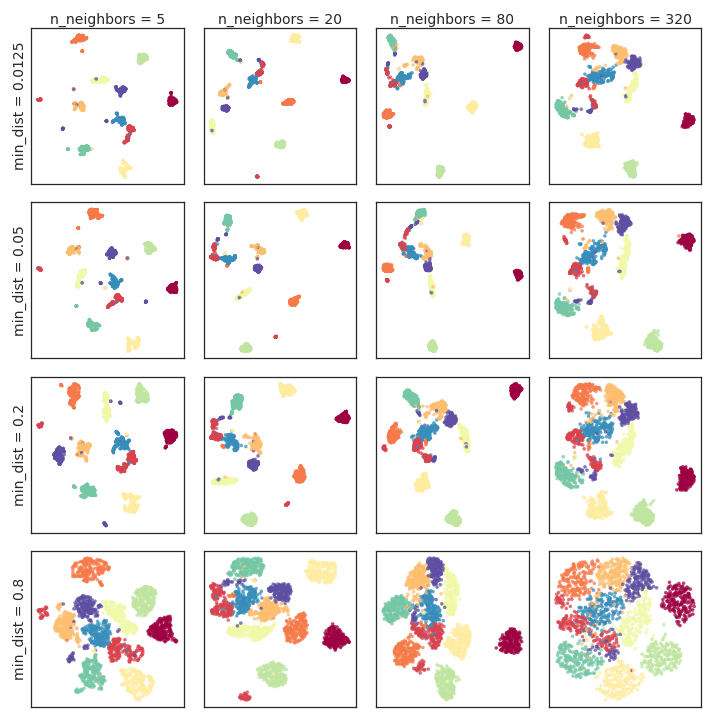}
    \caption{Variation of UMAP hyperparameters $n$ and min-dist result in different embeddings. The data is the PenDigits dataset, where each point is an 8x8 grayscale image of a hand-written digit.}
    \label{fig:hyperparameters-pendigits}
\end{figure}

In Figure \ref{fig:hyperparameters-pendigits} we provides examples of the same hyperparamter choices as Figure \ref{fig:hyperparameters}, but for the PenDigits dataset\footnote{See Section \ref{experiments} for a description of the PenDigits dataset}. In this case we expect small to medium $n$ values to be most effective, since there is significant cluster structure naturally present in the data. The min-dist parameter expands out tightly clustered groups, allowing more of the internal structure of densely packed clusters to be seen.

\begin{figure}
    \centering
    \includegraphics[width=0.9\textwidth]{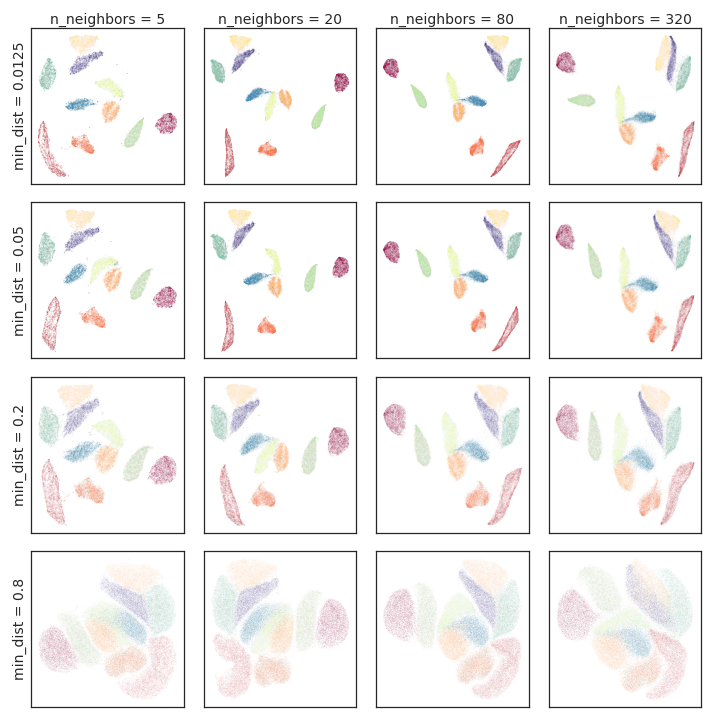}
    \caption{Variation of UMAP hyperparameters $n$ and min-dist result in different embeddings. The data is the MNIST dataset, where each point is an 28x28 grayscale image of a hand-written digit.}
    \label{fig:hyperparameters-mnist}
\end{figure}

Finally, in Figure \ref{fig:hyperparameters-mnist} we provide an equivalent example of hyperparameter choices for the MNIST dataset\footnote{See section \ref{experiments} for details on the MNIST dataset}. Again, since this dataset is expected to have signifcant cluster structure we expect medium sized values of $n$ to be most effective. We note that large values of min-dist result in the distinct clusters being compressed together, making the distinctions between the clusters less clear.

\section{Practical Efficacy}\label{experiments}

While the strong mathematical foundations of UMAP were the motivation for its development, the algorithm must ultimately be judged by its practical efficacy. In this section we examine the fidelity and performance of low dimensional embeddings of multiple diverse real world data sets under UMAP. The following datasets were considered:

\noindent\textbf{Pen digits} \cite{alpaydin1998pen, sklearn_api} is a set of 1797 grayscale images of digits entered using a digitiser tablet. Each image is an 8x8 image which we treat as a single 64 dimensional vector, assumed to be in Euclidean vector space.
\\
\textbf{COIL 20} \cite{COIL20}
is a set of 1440 greyscale images consisting of 20 objects under 72 different rotations spanning 360 degrees.  Each image is a 128x128 image which we treat as a single 16384 dimensional vector for the purposes of computing distance between images. \\
\textbf{COIL 100} \cite{COIL100}
is a set of 7200 colour images consisting of 100 objects under 72 different rotations spanning 360 degrees.  Each image consists of 3 128x128 intensity matrices (one for each color channel).  We treat this as a single 49152 dimensional vector for the purposes of computing distance between images. \\
\textbf{Mouse scRNA-seq} \cite{campbell2017molecular} is profiled gene expression data for 20,921 cells from an adult mouse.  Each sample consists of a vector of 26,774 measurements.\\
\textbf{Statlog (Shuttle)} \cite{UCI} is a NASA dataset consisting of various data associated to the positions of radiators in the space shuttle, including a timestamp. The dataset has 58000 points in a 9 dimensional feature space. \\
\textbf{MNIST} \cite{mnistlecun} is a dataset of 28x28 pixel grayscale images of handwritten digits. There are 10 digit classes (0 through 9) and 70000 total images. This is treated as 70000 different 784 dimensional vectors.\\
\textbf{F-MNIST} \cite{xiao2017} or Fashion MNIST is a dataset of 28x28 pixel grayscale images of fashion items (clothing, footwear and bags). There are 10 classes and 70000 total images. As with MNIST this is treated as 70000 different 784 dimensional vectors.\\
\textbf{Flow cytometry} \cite{spidlen2012flowrepository, brodie2013omip} is a dataset of flow cytometry measurements of CDT4 cells comprised of 1,000,000 samples, each with 17 measurements.
\\
\textbf{GoogleNews word vectors} \cite{mikolov2013distributed}
is a dataset of 3 million words and phrases derived from a sample of Google News documents and embedded into a 300 dimensional space via word2vec.\\

For all the datasets except GoogleNews we use Euclidean distance between vectors. For GoogleNews, as per \cite{mikolov2013distributed}, we use cosine distance (or angular distance in t-SNE which does support non-metric distances, in contrast to UMAP).

\subsection{Qualitative Comparison of Multiple Algorithms}\label{subsec:qual-comparison}

We compare a number of algorithms--UMAP, t-SNE \cite{vanDerMaaten2008,van2014accelerating}, LargeVis \cite{tang2016visualizing}, Laplacian Eigenmaps \cite{belkin2003laplacian}, and Principal Component Analysis \cite{hotelling1933analysis}--on the COIL20 \cite{COIL20}, MNIST \cite{mnistlecun}, Fashion-MNIST \cite{xiao2017}, and GoogleNews \cite{mikolov2013distributed} datasets. The Isomap algorithm was also tested, but failed to complete in any reasonable time for any of the datasets larger than COIL20.

The Multicore t-SNE package \cite{Ulyanov2016} was used for t-SNE. The reference implementation \cite{LFerry2016} was used for LargeVis. The scikit-learn \cite{sklearn_api} implementations were used for Laplacian Eigenmaps and PCA. Where possible we attempted to tune parameters for each algorithm to give good embeddings. 

\begin{figure}
    \centering
    \includegraphics[width=\textwidth]{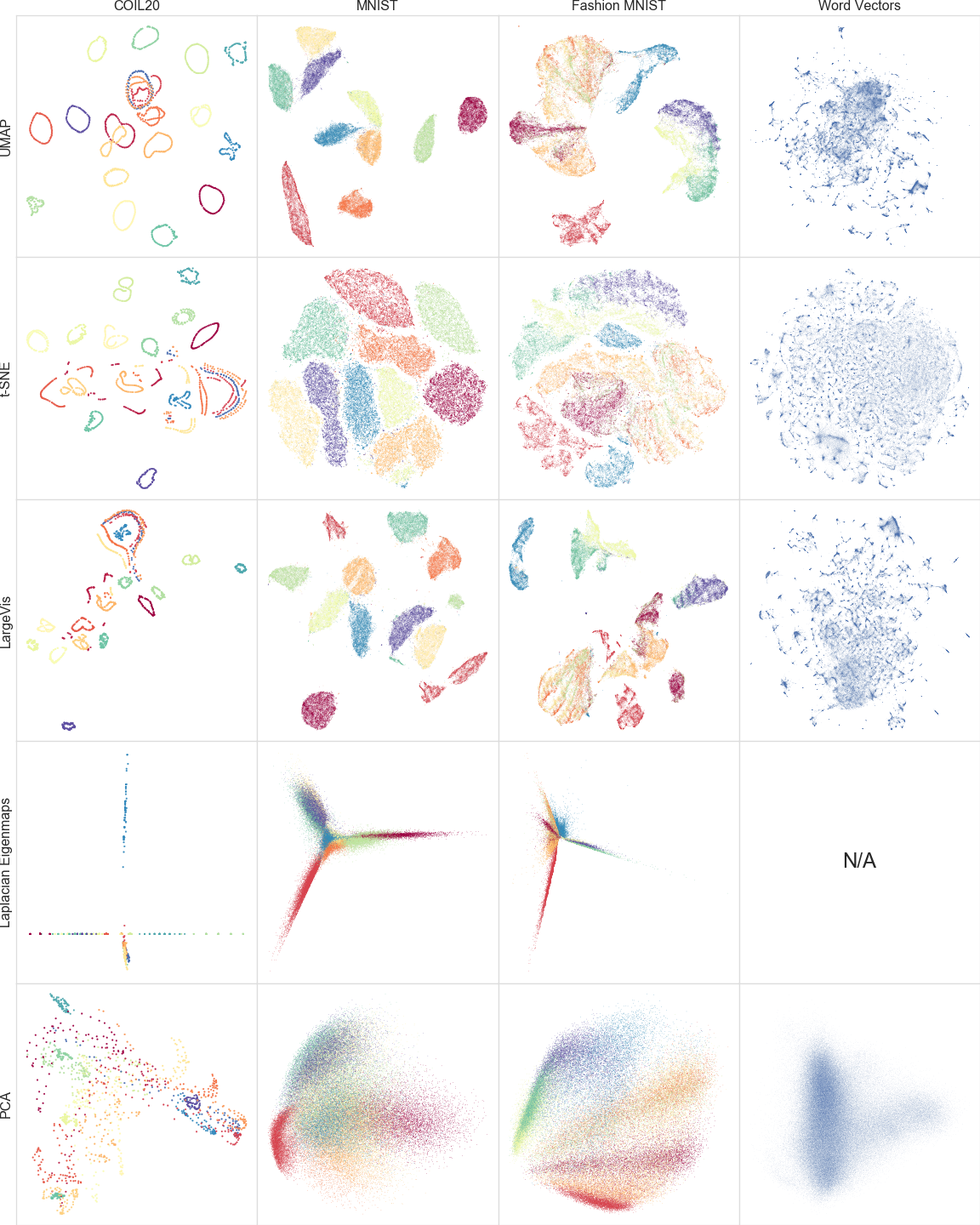}
    \caption{A comparison of several dimension reduction algorithms. We note that UMAP successfully reflects much of the large scale global structure that is well represented by Laplacian Eigenmaps and PCA (particularly for MNIST and Fashion-MNIST), while also preserving the local fine structure similar to t-SNE and LargeVis.}
    \label{fig:comparison}
\end{figure}


Historically t-SNE and LargeVis have offered a dramatic improvement in finding and preserving local structure in the data.  This can be seen qualitatively by comparing their embeddings to those generated by Laplacian Eigenmaps and PCA in Figure \ref{fig:comparison}.   We claim that the quality of embeddings produced by UMAP is comparable to t-SNE when reducing to two or three dimensions. For example, Figure \ref{fig:comparison} shows both UMAP and t-SNE embeddings of the COIL20, MNIST, Fashion MNIST, and Google News datasets. While the precise embeddings are different, UMAP distinguishes the same structures as t-SNE and LargeVis.

It can be argued that UMAP has captured more of the global and topological structure of the datasets than t-SNE \cite{Becht298430, wu2019comparison}. More of the loops in the COIL20 dataset are kept intact, including the intertwined loops. Similarly the global relationships among different digits in the MNIST digits dataset are more clearly captured with 1 (red) and 0 (dark red) at far corners of the embedding space, and 4,7,9 (yellow, sea-green, and violet) and 3,5,8 (orange, chartreuse, and blue) separated as distinct clumps of similar digits. In the Fashion MNIST dataset the distinction between clothing (dark red, yellow, orange, vermilion) and footwear (chartreuse, sea-green, and violet) is made more clear. Finally, while both t-SNE and UMAP capture groups of similar word vectors, the UMAP embedding arguably evidences a clearer global structure among the various word clusters.

\subsection{Quantitative Comparison of Multiple Algorithms}

We compare UMAP, t-SNE, LargeVis, Laplacian Eigenmaps and PCA embeddings with respect to the performance of a $k$-nearest neighbor classifier trained on the embedding space for a variety of datasets. The $k$-nearest neighbor classifier accuracy provides a clear quantitative measure of how well the embedding has preserved the important local structure of the dataset. By varying the hyper-parameter $k$ used in the training we can also consider how structure preservation varies under transition from purely local to non-local, to more global structure. The embeddings used for training the $k$NN classifier are for those datasets that come with defined training labels: PenDigits, COIL-20, Shuttle, MNIST, and Fashion-MNIST.

We divide the datasets into two classes: smaller datasets (PenDigits and COIL-20), for which a smaller range of $k$ values makes sense, and larger datasets, for which much larger values of $k$ are reasonable. For each of the small datasets a stratified 10-fold cross-validation was used to derive a set of 10 accuracy scores for each embedding. For the Shuttle dataset a 10-fold cross-validation was used due to constraints imposed by class sizes and the stratified sampling. For MNIST and Fashion-MNIST a 20-fold cross validation was used, producing 20 accuracy scores.

\begin{figure}[!hptb]
    \centering
    \includegraphics[width=\textwidth]{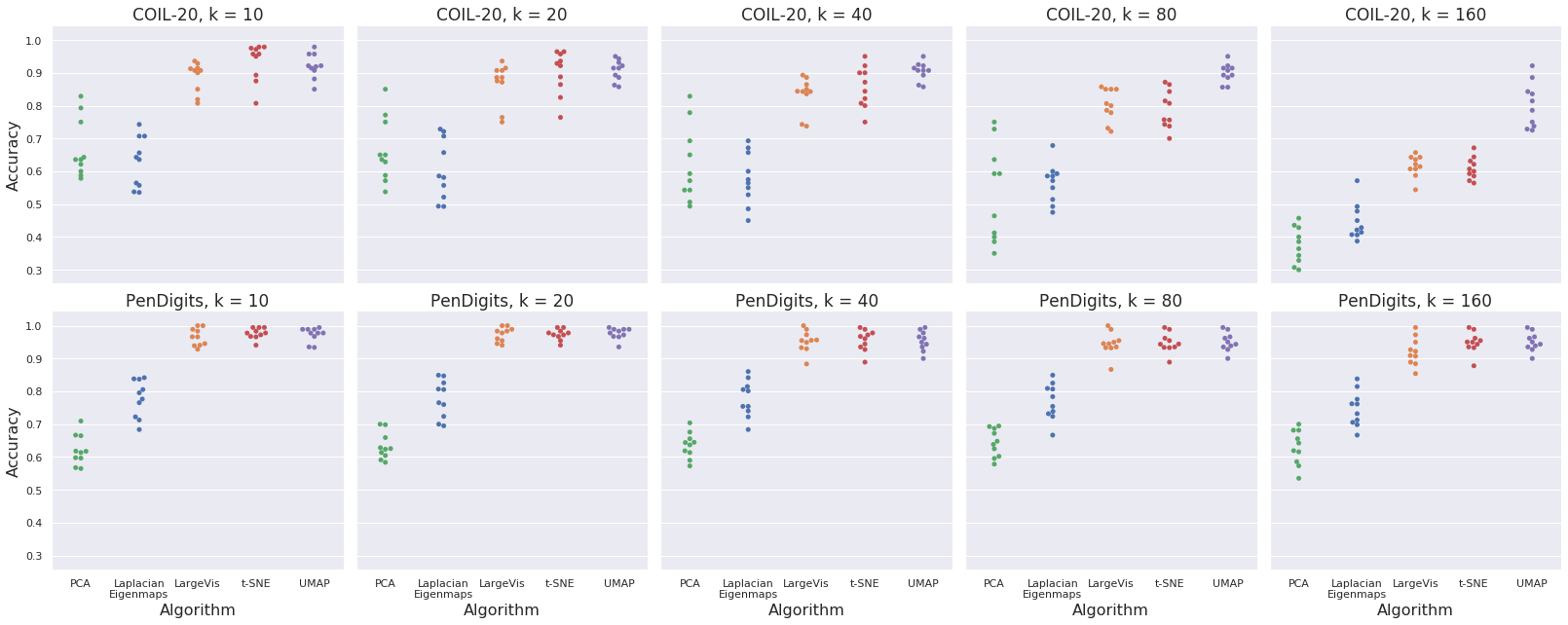}
    \caption{$k$NN Classifier accuracy for varying values of $k$ over the embedding spaces of COIL-20
    and PenDigits datasets. Accuracy scores are given for each fold of a 10-fold cross-validation for
    each of PCA, Laplacian Eigenmaps, LargeVis, t-SNE and UMAP. We note that UMAP produces competitive
    accuracy scores to t-SNE and LargeVis for most cases, and outperforms both t-SNE and LargeVis
    for larger $k$ values on COIL-20.}
    \label{fig:knn_crossval_small_data}
\end{figure}

In Table \ref{table:small-data-accuracy} we present the average accuracy across the 10-folds for the PenDigits and COIL-20 datasets. UMAP performs at least as well as t-SNE and LargeVis (given the confidence bounds on the accuracy) for $k$ in the range 10 to 40, but for larger $k$ values of 80 and 160 UMAP has significantly higher accuracy on COIL-20, and shows evidence of higher accuracy on PenDigits. Figure \ref{fig:knn_crossval_small_data} provides swarm plots of the accuracy results across the COIL-20 and PenDigits datasets.

\begin{figure}[!hptb]
    \centering
    \includegraphics[width=\textwidth]{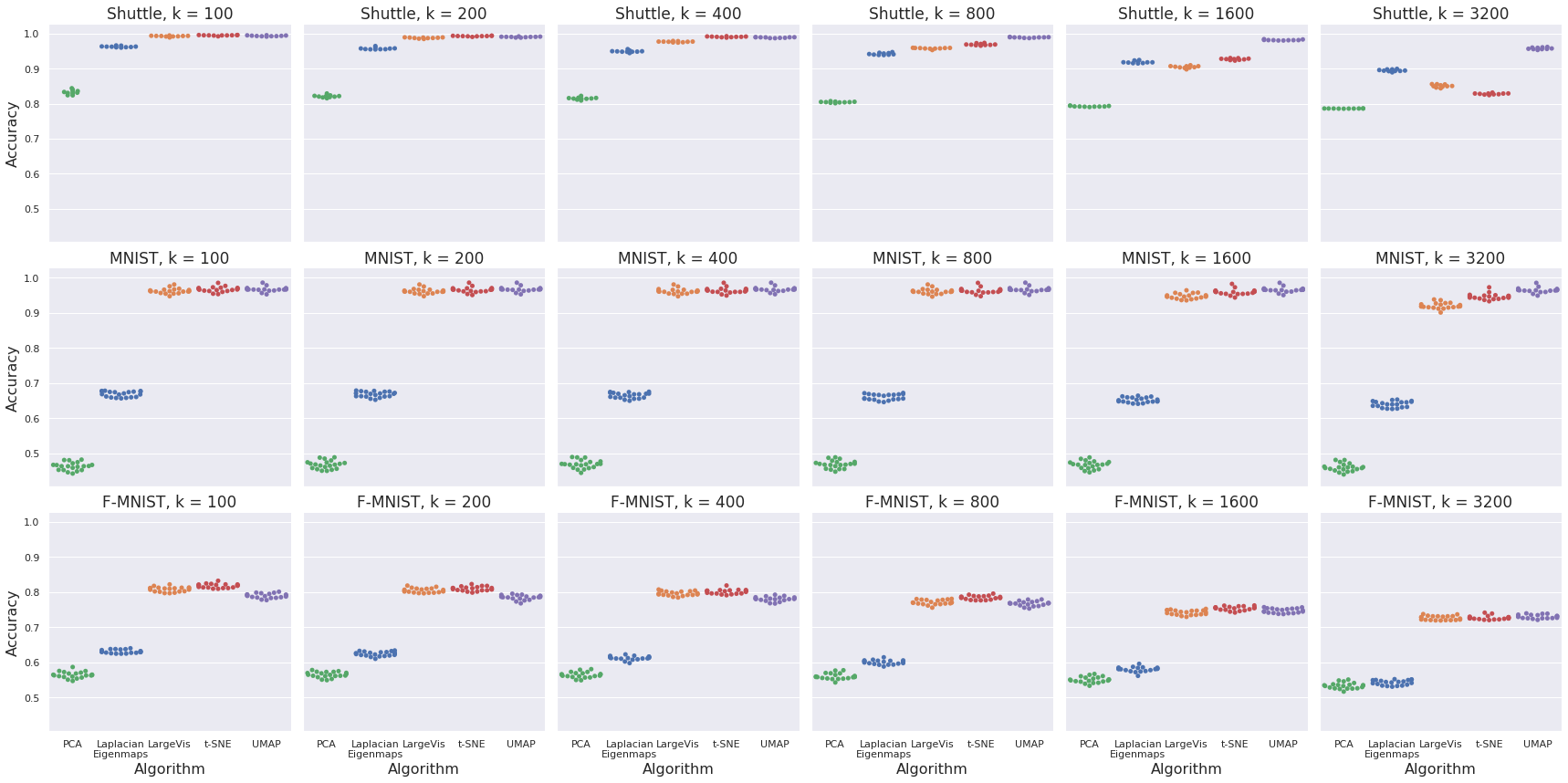}
    \caption{$k$NN Classifier accuracy for varying values of $k$ over the embedding spaces of Shuttle, MNIST and Fashion-MNIST datasets. Accuracy scores are given for each fold of a 10-fold cross-validation for
     Shuttle, and 20-fold cross-validation for MNIST and Fashion-MNIST, for each of PCA, Laplacian Eigenmaps, LargeVis, t-SNE and UMAP. UMAP performs better than the other algorithms for large $k$, particularly on the Shuttle dataset. For Fashion-MNIST UMAP provides slightly poorer accuracy than t-SNE and LargeVis at small scales, but is competitive at larger $k$ values.}
    \label{fig:knn_crossval_large_data}
\end{figure}

In Table \ref{table:large-data-accuracy} we present the average cross validation accuracy for the Shuttle, MNIST and Fashion-MNIST datasets. UMAP performs at least as well as t-SNE and LargeVis (given the confidence bounds on the accuracy) for $k$ in the range 100 to 400 on the Shuttle and MNIST datasets (but notably underperforms on the Fashion-MNIST dataset), but for larger $k$ values of 800 and 3200 UMAP has significantly higher accuracy on the Shuttle dataset, and shows evidence of higher accuracy on MNIST. For $k$ values of 1600 and 3200 UMAP establishes comparable performance on Fashion-MNIST. Figure \ref{fig:knn_crossval_large_data} provides swarm plots of the accuracy results across the Shuttle and MNIST and Fashion-MNIST datasets.

\begin{table}[!hptb]
\centering
\begin{tabular}{ll|rrrrr}
\toprule
\textbf{} & \textbf{k} & \textbf{t-SNE} & \textbf{UMAP} & \textbf{LargeVis} & \textbf{Eigenmaps} & \textbf{PCA} \\
\cline{2-7}
\parbox[t]{2mm}{\multirow{5}{*}{\rotatebox[origin=c]{90}{COIL-20}}}
          & 10 & \textbf{0.934} {\tiny ($\pm$ 0.115)}  & 0.921 {\tiny ($\pm$ 0.075)} & 0.888  {\tiny ($\pm$ 0.092)}& 0.629 {\tiny ($\pm$ 0.153)} & 0.667  {\tiny ($\pm$ 0.179)}   \\
          & 20 & 0.901  {\tiny ($\pm$ 0.133)}  & \textbf{0.907} {\tiny ($\pm$ 0.064)} & 0.870 {\tiny ($\pm$ 0.125)} & 0.605 {\tiny ($\pm$ 0.185)} & 0.663  {\tiny ($\pm$ 0.196)}   \\
          & 40 & 0.857  {\tiny ($\pm$ 0.125)}  & \textbf{0.904} {\tiny ($\pm$ 0.056)} & 0.833 {\tiny ($\pm$ 0.106)} & 0.578 {\tiny ($\pm$ 0.159)} & 0.620  {\tiny ($\pm$ 0.230)}   \\
          & 80 & 0.789  {\tiny ($\pm$ 0.118)}  & \textbf{0.899} {\tiny ($\pm$ 0.058)} & 0.803 {\tiny ($\pm$ 0.100)} & 0.565 {\tiny ($\pm$ 0.119)} & 0.531  {\tiny ($\pm$ 0.294)}   \\
          & 160 & 0.609  {\tiny ($\pm$ 0.067)} & \textbf{0.803} {\tiny ($\pm$ 0.138)} & 0.616 {\tiny ($\pm$ 0.066)} & 0.446 {\tiny ($\pm$ 0.110)} & 0.375 {\tiny ($\pm$ 0.111)}    \\
\cline{2-7}
\parbox[t]{2mm}{\multirow{5}{*}{\rotatebox[origin=c]{90}{PenDigits}}}
    & 10	& \textbf{0.977}	{\tiny ($\pm$ 0.033)}& 0.973 {\tiny ($\pm$ 0.044)}	& 0.966	{\tiny ($\pm$ 0.053)}&  0.778	{\tiny ($\pm$ 0.113)}& 0.622 {\tiny ($\pm$ 0.092)} \\
    & 20	& 0.973	{\tiny ($\pm$ 0.033)}& \textbf{0.976} {\tiny ($\pm$ 0.035)}	& 0.973	{\tiny ($\pm$ 0.044)}&  0.778	{\tiny ($\pm$ 0.116)}& 0.633 {\tiny ($\pm$ 0.082)} \\
    & 40	& 0.956	{\tiny ($\pm$ 0.064)}& 0.954 {\tiny ($\pm$ 0.060)}  & \textbf{0.959}	{\tiny ($\pm$ 0.066)}&  0.778	{\tiny ($\pm$ 0.112)}& 0.636 {\tiny ($\pm$ 0.078)} \\
    & 80	& 0.948	{\tiny ($\pm$ 0.060)}& \textbf{0.951} {\tiny ($\pm$ 0.072)}  & 0.949	{\tiny ($\pm$ 0.072)}&  0.767	{\tiny ($\pm$ 0.111)}& 0.643 {\tiny ($\pm$ 0.085)} \\
    & 160	& 0.949	{\tiny ($\pm$ 0.065)}& \textbf{0.951} {\tiny ($\pm$ 0.085)}	& 0.921	{\tiny ($\pm$ 0.085)}&  0.747	{\tiny ($\pm$ 0.108)}& 0.629 {\tiny ($\pm$ 0.107)} \\
\bottomrule
\end{tabular}
\caption{$k$NN Classifier accuracy for varying values of $k$ over the embedding spaces of COIL-20
    and PenDigits datasets. Average accuracy scores are given over a 10-fold cross-validation for
    each of PCA, Laplacian Eigenmaps, LargeVis, t-SNE and UMAP.}\label{table:small-data-accuracy}
\end{table}

\begin{table}[!hptb]
\centering
\begin{tabular}{ll|rrrrr}
\toprule
\textbf{} & \textbf{k} & \textbf{t-SNE} & \textbf{UMAP} & \textbf{LargeVis} & \textbf{Eigenmaps} & \textbf{PCA} \\
\cline{2-7}
\parbox[t]{2mm}{\multirow{6}{*}{\rotatebox[origin=c]{90}{Shuttle}}}
&100	&\textbf{0.994}	{\tiny ($\pm$ 0.002)}&0.993	{\tiny ($\pm$ 0.002)}&0.992	{\tiny ($\pm$ 0.003)}&0.962	{\tiny ($\pm$ 0.004)}&0.833  {\tiny ($\pm$ 0.013)}\\
&200	&\textbf{0.992}	{\tiny ($\pm$ 0.002)}&0.990	{\tiny ($\pm$ 0.002)}&0.987	{\tiny ($\pm$ 0.003)}&0.957	{\tiny ($\pm$ 0.006)}&0.821  {\tiny ($\pm$ 0.007)}\\
&400	&\textbf{0.990}	{\tiny ($\pm$ 0.002)}&0.988	{\tiny ($\pm$ 0.002)}&0.976	{\tiny ($\pm$ 0.003)}&0.949	{\tiny ($\pm$ 0.006)}&0.815  {\tiny ($\pm$ 0.007)}\\
&800	&0.969	{\tiny ($\pm$ 0.005)}&\textbf{0.988}	{\tiny ($\pm$ 0.002)}&0.957	{\tiny ($\pm$ 0.004)}&0.942	{\tiny ($\pm$ 0.006)}&0.804  {\tiny ($\pm$ 0.003)}\\
&1600	&0.927	{\tiny ($\pm$ 0.005)}&\textbf{0.981}	{\tiny ($\pm$ 0.002)}&0.904	{\tiny ($\pm$ 0.007)}&0.918	{\tiny ($\pm$ 0.006)}&0.792  {\tiny ($\pm$ 0.003)}\\
&3200	&0.828	{\tiny ($\pm$ 0.004)}&\textbf{0.957}	{\tiny ($\pm$ 0.005)}&0.850	{\tiny ($\pm$ 0.008)}&0.895	{\tiny ($\pm$ 0.006)}&0.786  {\tiny ($\pm$ 0.001)}\\
\cline{2-7}
\parbox[t]{2mm}{\multirow{6}{*}{\rotatebox[origin=c]{90}{MNIST}}}
&100	&\textbf{0.967}	{\tiny ($\pm$ 0.015)}&\textbf{0.967}	{\tiny ($\pm$ 0.014)}&0.962	{\tiny ($\pm$ 0.015)}&0.668	{\tiny ($\pm$ 0.016)}&0.462  {\tiny ($\pm$ 0.023)}\\
&200	&0.966	{\tiny ($\pm$ 0.015)}&\textbf{0.967}	{\tiny ($\pm$ 0.014)}&0.962	{\tiny ($\pm$ 0.015)}&0.667	{\tiny ($\pm$ 0.016)}&0.467  {\tiny ($\pm$ 0.023)}\\
&400	&0.964	{\tiny ($\pm$ 0.015)}&\textbf{0.967}	{\tiny ($\pm$ 0.014)}&0.961	{\tiny ($\pm$ 0.015)}&0.664	{\tiny ($\pm$ 0.016)}&0.468  {\tiny ($\pm$ 0.024)}\\
&800	&0.963	{\tiny ($\pm$ 0.016)}&\textbf{0.967}	{\tiny ($\pm$ 0.014)}&0.961	{\tiny ($\pm$ 0.015)}&0.660	{\tiny ($\pm$ 0.017)}&0.468  {\tiny ($\pm$ 0.023)}\\
&1600	&0.959	{\tiny ($\pm$ 0.016)}&\textbf{0.966}	{\tiny ($\pm$ 0.014)}&0.947	{\tiny ($\pm$ 0.015)}&0.651	{\tiny ($\pm$ 0.014)}&0.467  {\tiny ($\pm$ 0.0233)}\\
&3200	&0.946	{\tiny ($\pm$ 0.017)}&\textbf{0.964}	{\tiny ($\pm$ 0.014)}&0.920	{\tiny ($\pm$ 0.017)}&0.639	{\tiny ($\pm$ 0.017)}&0.459  {\tiny ($\pm$ 0.022)}\\
\cline{2-7}
\parbox[t]{2mm}{\multirow{6}{*}{\rotatebox[origin=c]{90}{Fashion-MNIST}}}
&100	&\textbf{0.818}	{\tiny ($\pm$ 0.012)}&0.790	{\tiny ($\pm$ 0.013)}&0.808	{\tiny ($\pm$ 0.014)}&0.631	{\tiny ($\pm$ 0.010)}&0.564  {\tiny ($\pm$ 0.018)}\\
&200	&\textbf{0.810}	{\tiny ($\pm$ 0.013)}&0.785	{\tiny ($\pm$ 0.014)}&0.805	{\tiny ($\pm$ 0.013)}&0.624	{\tiny ($\pm$ 0.013)}&0.565  {\tiny ($\pm$ 0.016)}\\
&400	&\textbf{0.801}	{\tiny ($\pm$ 0.013)}&0.780	{\tiny ($\pm$ 0.013)}&0.796	{\tiny ($\pm$ 0.013)}&0.612	{\tiny ($\pm$ 0.011)}&0.564  {\tiny ($\pm$ 0.017)}\\
&800	&\textbf{0.784}	{\tiny ($\pm$ 0.011)}&0.767	{\tiny ($\pm$ 0.014)}&0.771	{\tiny ($\pm$ 0.014)}&0.600	{\tiny ($\pm$ 0.012)}&0.560  {\tiny ($\pm$ 0.017)}\\
&1600	&\textbf{0.754}	{\tiny ($\pm$ 0.011)}&0.747	{\tiny ($\pm$ 0.013)}&0.742	{\tiny ($\pm$ 0.013)}&0.580	{\tiny ($\pm$ 0.014)}&0.550  {\tiny ($\pm$ 0.017)}\\
&3200	&0.727	{\tiny ($\pm$ 0.011)}&\textbf{0.730}	{\tiny ($\pm$ 0.011)}&0.726	{\tiny ($\pm$ 0.012)}&0.542	{\tiny ($\pm$ 0.014)}&0.533  {\tiny ($\pm$ 0.017)}\\
\bottomrule
\end{tabular}
\caption{$k$NN Classifier accuracy for varying values of $k$ over the embedding spaces of Shuttle, MNIST
    and Fashion-MNIST datasets. Average accuracy scores are given over a 10-fold or 20-fold cross-validation for each of PCA, Laplacian Eigenmaps, LargeVis, t-SNE and UMAP.}\label{table:large-data-accuracy}
\end{table}

As evidenced by this comparison UMAP provides largely comparable perfomance in embedding quality to t-SNE and LargeVis at local scales, but performs markedly better than t-SNE or LargeVis at non-local scales. This bears out the visual qualitative assessment provided in Subsection \ref{subsec:qual-comparison}.

\subsection{Embedding Stability}

Since UMAP makes use of both stochastic approximate nearest neighbor search, and stochastic gradient descent with negative sampling for optimization, the resulting embedding is necessarily different from run to run, and under sub-sampling of the data. This is potentially a concern for a variety of uses cases, so establishing some measure of how stable UMAP embeddings are, particularly under sub-sampling, is of interest. In this subsection we compare the stability under subsampling of UMAP, LargeVis and t-SNE (the three stochastic dimension reduction techniques considered).

\begin{figure}
    \centering
    \begin{subfigure}{0.48\textwidth}
        \includegraphics[width=\textwidth]{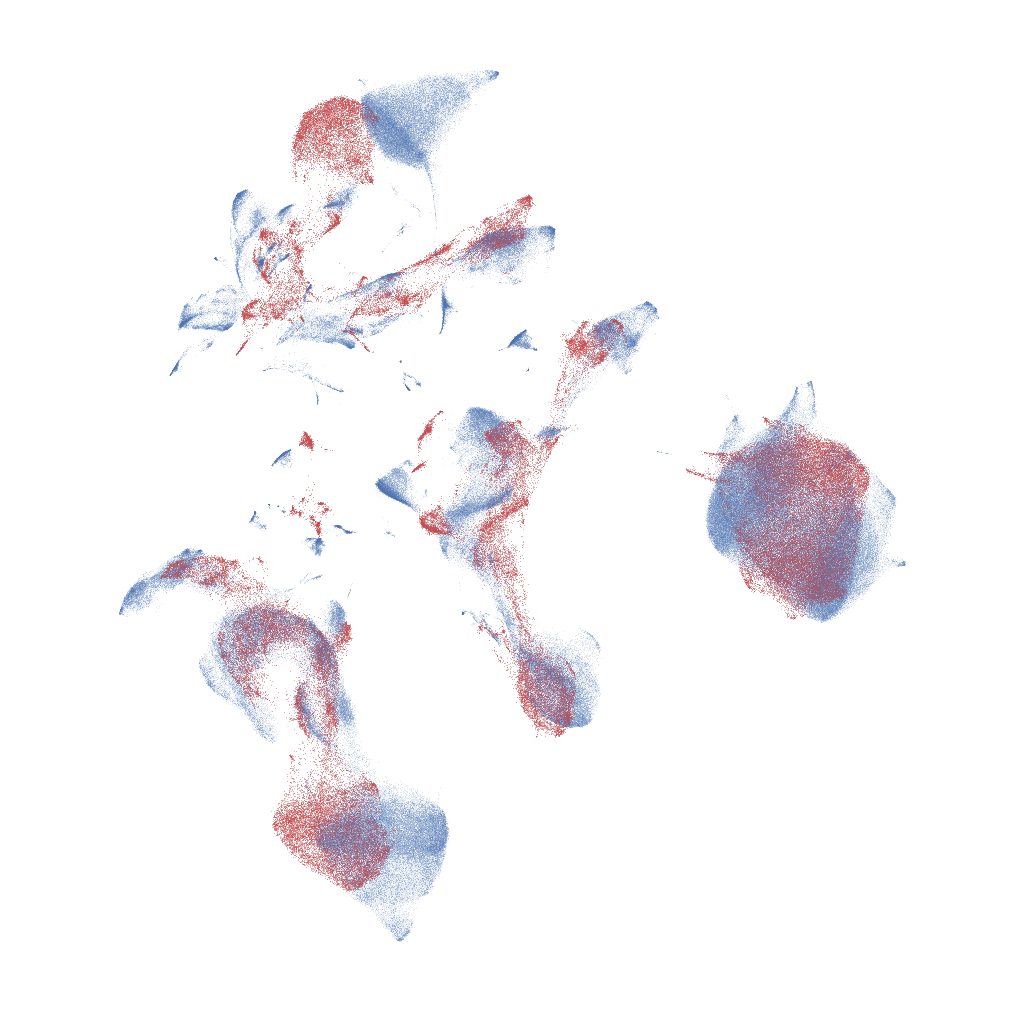}
        \caption{UMAP}
    \end{subfigure}
    \begin{subfigure}{0.48\textwidth}
        \includegraphics[width=\textwidth]{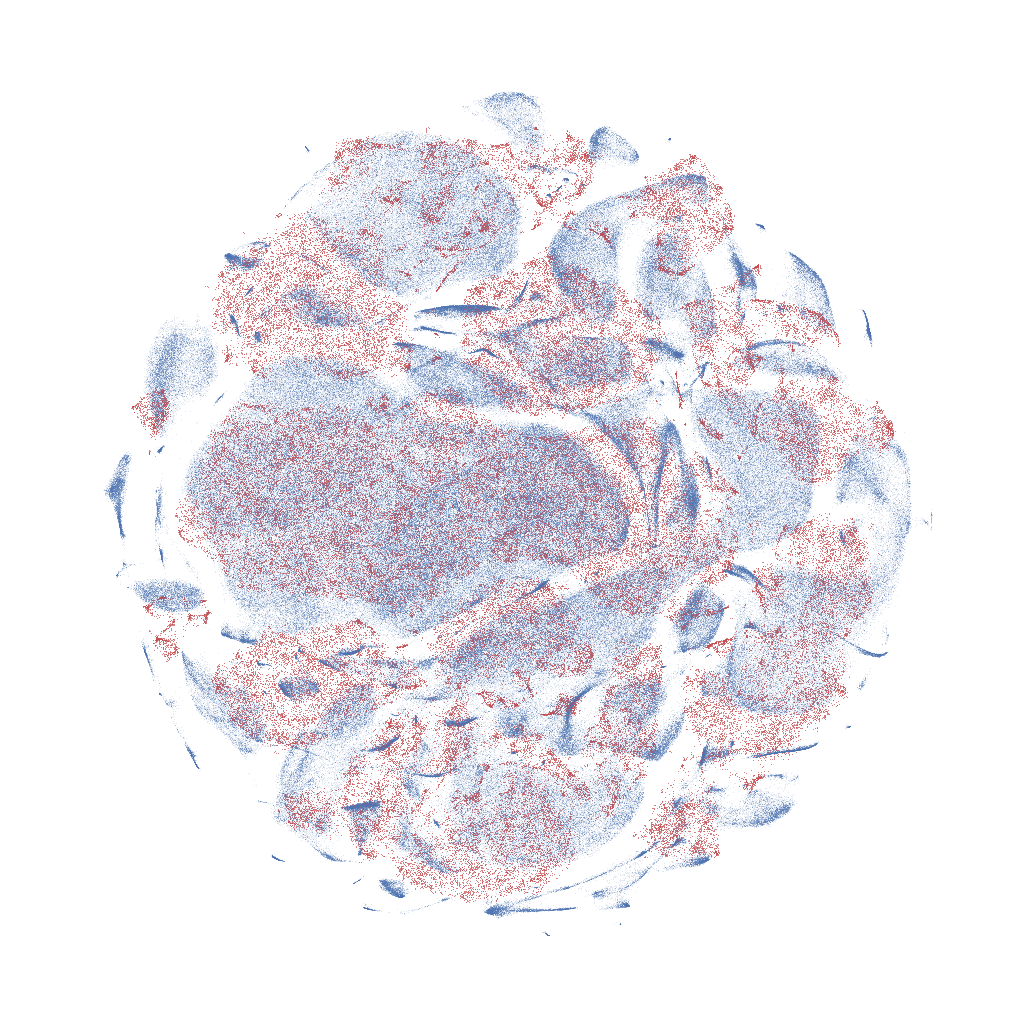}
        \caption{t-SNE}
    \end{subfigure}
    \caption{Procrustes based alignment of a 10\% subsample (red) against the full dataset (blue) for the flow cytometry dataset for both UMAP and t-SNE.}
    \label{fig:procrustes-alignment}
\end{figure}

To measure the stability of an embedding we make use of the normalized Procrustes distance to measure the distance between two potentially comparable distributions. Given two datasets $X = \{x_1, \ldots, x_N\}$ and $Y = \{y_1, \ldots, y_N\}$ such that $x_i$ corresponds to $y_i$, we can define the Procustes distance between the datasets $d_P(X, Y)$ in the following manner. Determine $Y' = \{{y_1}',\ldots, {y_N}'\}$ the optimal translation, uniform scaling, and rotation of $Y$ that minimizes the squared error $\sum_{i=1}^N (x_i - {y_i}')^2$, and define 
\[
d_P(X, Y) = \sqrt{\sum_{i=1}^N (x_i - {y_i}')^2}.
\]
Since any measure that makes use of distances in the embedding space is potentially sensitive to the extent or scale of the embedding, we normalize the data before computing the Procrustes distance by dividing by the average norm of the embedded dataset. In Figure \ref{fig:procrustes-alignment} we visualize the results of using Procrustes alignment of embedding of sub-samples for both UMAP and t-SNE, demonstrating how Procrustes distance can measure the stability of the overall structure of the embedding. 

Given a measure of distance between different embeddings we can examine stability under sub-sampling by considering the normalized Procrustes distance between the embedding of a sub-sample, and the corresponding sub-sample of an embedding of the full dataset. As the size of the sub-sample increases the average distance per point between the sub-sampled embeddings should decrease, potentially toward some asymptote of maximal agreement under repeated runs. Ideally this asymptotic value would be zero error, but for stochastic embeddings such as UMAP and t-SNE this is not achievable.

\begin{figure}
    \centering
    \includegraphics[width=\textwidth]{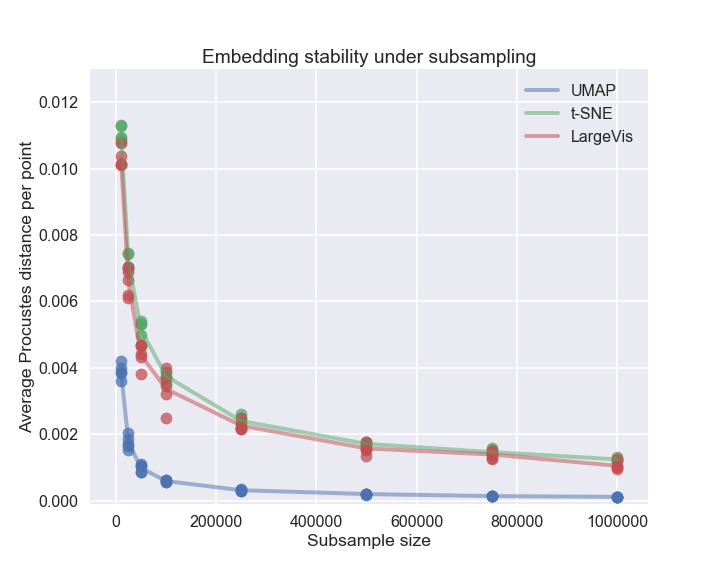}
    \caption{Comparison of average Procustes distance per point for t-SNE, LargeVis and UMAP over a variety of sizes of subsamples from the full  Flow Cytometry dataset. UMAP sub-sample embeddings are very close to the full embedding even for subsamples of 5\% of the full dataset, outperforming the results of t-SNE and LargeVis even when they use the full Flow Cytometry dataset.}
    \label{fig:procustes-results}
\end{figure}

We performed an empirical comparison of algorithms with respect to stability using the Flow Cytometry dataset due its large size, interesting structure, and low ambient dimensionality (aiding runtime performance for t-SNE). We note that for a dataset this large we found it necessary to increase the default \verb+n_iter+ value for t-SNE from 1000 to 1500 to ensure better convergence. While this had an impact on the runtime, it significantly improved the Procrustes distance results by providing more stable and consistent embeddings. Figure \ref{fig:procustes-results} provides a comparison between UMAP and t-SNE, demonstrating that UMAP has signifcantly more stable results than t-SNE. In particular, after sub-sampling on 5\% of the million data points, the per point error for UMAP was already below any value achieved by t-SNE. 

\subsection{Computational Performance Comparisons}\label{Computation Performance}

Benchmarks against the real world datasets were performed on a Macbook Pro with a 3.1 GHz Intel Core i7 and 8GB of RAM for Table \ref{tab:performance}, and on a server with Intel Xeon E5-2697v4 processors and 512GB of RAM for the large scale benchmarking in Subsections \ref{embed_dim_scaling}, \ref{ambient_dim_scaling}, and \ref{data_scaling}.

For t-SNE we chose MulticoreTSNE \cite{Ulyanov2016}, which we believe to be the fastest extant implementation of Barnes-Hut t-SNE at this time, even when run in single core mode. It should be noted that MulticoreTSNE is a heavily optimized implementation written in C++ based on Van der Maaten's \verb+bhtsne+ \cite{van2014accelerating} code.

As a fast alternative approach to t-SNE we also consider the FIt-SNE algorithm \cite{linderman2017efficient}. We used the reference implementation \cite{FItSNE2018}, which, like MulticoreTNSE is an optimized C++ implementation. We also note that FIt-SNE makes use of multiple cores.

LargeVis \cite{tang2016visualizing} was benchmarked using the reference implementation \cite{LFerry2016}. It was run with default parameters including use of 8 threads on the 4-core machine. The only exceptions were small datasets where we explicitly set the \texttt{-samples} parameter to \verb+n_samples+/100 as per the recommended values in the documentation of the reference implementation.

The Isomap \cite{tenenbaumISOMAP} and Laplacian Eigenmaps \cite{belkin2003laplacian} implementations in scikit-learn \cite{sklearn_api} were used. We suspect the Laplacian eigenmaps implementation may not be well optimized for large datasets but did not find a better performing implementation that provided comparable quality results. Isomap failed to complete for the Shuttle, Fashion-MNIST, MNIST and GoogleNews datasets, while Laplacian Eigenmaps failed to run for the GoogleNews dataset.

To allow a broader range of algorithms to run some of the datasets where subsampled or had their dimension reduced by PCA.  The Flow Cytometry dataset was benchmarked on a 10\% sample and the GoogleNews was subsampled down to 200,000 data points.  Finally, the Mouse scRNA dataset was reduced to 1,000 dimensions via PCA.

Timing were performed for the COIL20 \cite{COIL20}, COIL100 \cite{COIL100}, Shuttle \cite{UCI}, MNIST \cite{mnistlecun}, Fashion-MNIST \cite{xiao2017}, and GoogleNews \cite{mikolov2013distributed} datasets. Results can be seen in Table \ref{tab:performance}. UMAP consistently performs faster than any of the other algorithms aside from on the very small Pendigits dataset, where Laplacian Eigenmaps and Isomap have a small edge.



\begin{table}[!hptb]
    \centering
    \begin{tabular}{rrrrrrr}
    \toprule
      & {\bf UMAP} & {\bf FIt-SNE} & {\bf t-SNE} & {\bf LargeVis} & {\bf Eigenmaps} & {\bf Isomap} \\
    \hline
    {\bf Pen Digits}& 9s & 48s & 17s & 20s & \textbf{2s} & \textbf{2s}\\
    {\small (1797x64)}& & & & & & \\[10pt]
    {\bf COIL20} & \textbf{12s} & 75s & 22s & 82s & 47s & 58s \\
    {\small (1440x16384)}& & & & & & \\[10pt]
    {\bf COIL100} & \textbf{85s} & 2681s & 810s & 3197s & 3268s & 3210s \\
    {\small (7200x49152)}& & & & & & \\[10pt]
    {\bf scRNA} & \textbf{28s} & 131s & 258s & 377s & 470s & 923s \\
    {\small (21086x1000)}& & & & & & \\[10pt]
    {\bf Shuttle} & \textbf{94s} & 108s & 714s & 615s & 133s &  -- \\
    {\small (58000x9)}& & & & & & \\[10pt]
    {\bf MNIST} & \textbf{87s} & 292s & 1450s & 1298s & 40709s & -- \\
    {\small (70000x784)}& & & & & & \\[10pt]
    {\bf F-MNIST} & \textbf{65s} & 278s & 934s & 1173s & 6356s & -- \\
    {\small (70000x784)}& & & & & & \\[10pt]
    {\bf Flow} & \textbf{102s} & 164s & 1135s & 1127s & 30654s & -- \\
    {\small (100000x17)}& & & & & & \\[10pt]
    {\bf Google News} & \textbf{361s} & 652s & 16906s & 5392s & -- & -- \\
    {\small (200000x300)}& & & & & & \\[10pt]
    \bottomrule
    \end{tabular}
    \caption{Runtime of several dimension reduction algorithms on various datasets. To allow a broader range of algorithms to run some of the datasets where subsampled or had their dimension reduced by PCA.  The Flow Cytometry dataset was benchmarked on a 10\% sample and the GoogleNews was subsampled down to 200,000 data points.  Finally, the Mouse scRNA dataset was reduced to 1,000 dimensions via PCA. The fastest runtime for each dataset has been bolded.}
    \label{tab:performance}

\end{table}

\subsubsection{Scaling with Embedding Dimension}\label{embed_dim_scaling}

\begin{figure}[!hbtp]
    \centering
    \begin{subfigure}[t]{0.45\textwidth}
        \includegraphics[width=\textwidth]{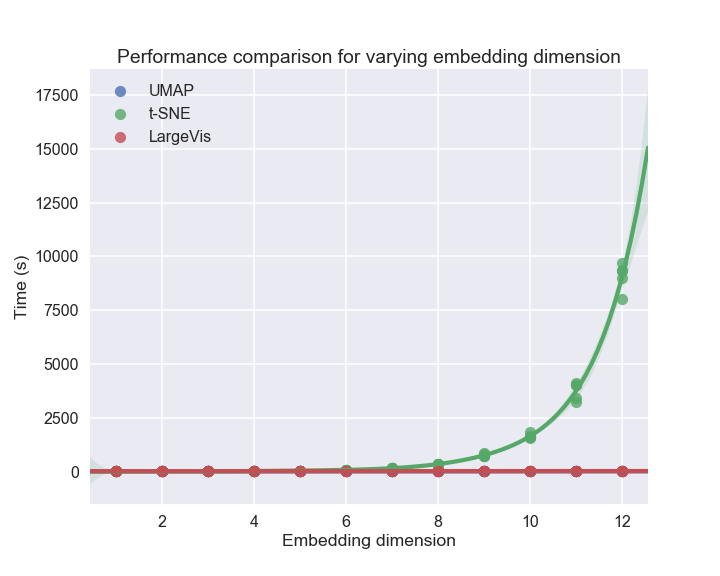}
        \caption{A comparison of run time for UMAP, t-SNE and LargeVis with respect to embedding dimension on the Pen digits dataset.  We see that t-SNE scales worse than exponentially while UMAP and LargeVis scale linearly with a slope so slight to be undetectable at this scale.}
        \label{fig:large_scale_embedding_dim}
    \end{subfigure}
    \quad
    \begin{subfigure}[t]{0.45\textwidth}
        \includegraphics[width=\textwidth]{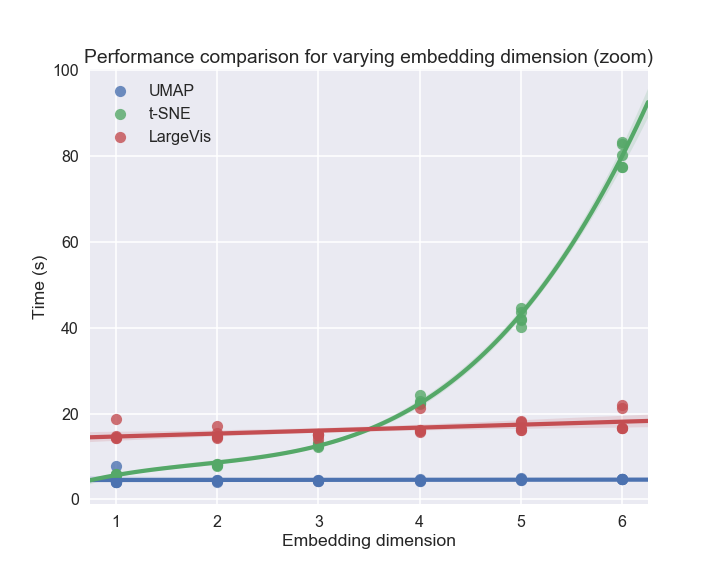}
        \caption{Detail of scaling for embedding dimension of six or less. We can see that UMAP and LargeVis are essentially flat. In practice they
        appear to scale linearly, but the slope is essentially undetectable at this scale.}
        \label{fig:small_scale_embedding_dim}
    \end{subfigure}
    \caption{Scaling performance with respect to embedding dimension of UMAP, t-SNE and LargeVis on the Pen digits dataset.}
    \label{fig:scale_embedding_dim_umap-tsne}
\end{figure}

UMAP is significantly more performant than t-SNE\footnote{Comparisons were performed against MulticoreTSNE as the current implementation of FIt-SNE does not support embedding into any dimension larger than 2.} when embedding into dimensions larger than 2. This is particularly important when the intention is to use the low dimensional representation for further machine learning tasks such as clustering or anomaly detection rather than merely for visualization. The computation performance of UMAP  is far more efficient than t-SNE, even for very small embedding dimensions of 6 or 8 (see Figure \ref{fig:scale_embedding_dim_umap-tsne}). This is largely due to the fact that UMAP does not require global normalisation (since it represents data as a fuzzy topological structure rather than as a probability distribution). This allows the algorithm to work without the need for space trees ---such as the quad-trees and oct-trees that t-SNE uses \cite{van2014accelerating}---. Such space trees scale exponentially in dimension, resulting in t-SNE's relatively poor scaling with respect to embedding dimension. By contrast, we see that UMAP consistently scales well in embedding dimension, making the algorithm practical for a wider range of applications beyond visualization.

\subsubsection{Scaling with Ambient Dimension}\label{ambient_dim_scaling}

Through a combination of the local-connectivity constraint and the approximate nearest neighbor search, UMAP can perform effective dimension reduction even for very high dimensional data (see Figure \ref{fig:umap_3e7_density} for an example of UMAP operating directly on 1.8 million dimensional data). This stands in contrast to many other manifold learning techniques, including t-SNE and LargeVis, for which it is generally recommended to reduce the dimension with PCA before applying these techniques (see \cite{maaten2008visualizing} for example).

\begin{figure}
    \centering
    \includegraphics[width=\textwidth]{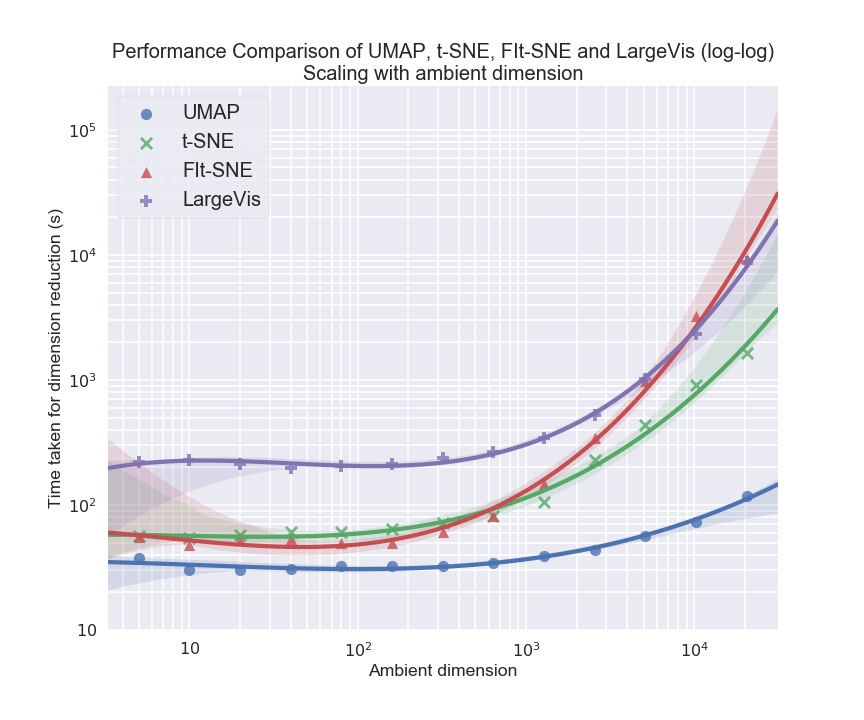}
    \caption{Runtime performance scaling of UMAP, t-SNE, FIt-SNE and Largevis with respect to the ambient dimension of the data. As the ambient dimension increases beyond a few thousand dimensions the computational cost of t-SNE, FIt-SNE, and LargeVis all increase dramatically, while UMAP continues to perform well into the tens-of-thousands of dimensions.}
    \label{fig:ambient-dim-scaling}
\end{figure}

To compare runtime performance scaling with respect to the ambient dimension of the data we chose to use the Mouse scRNA dataset, which is high dimensional, but is also amenable to the use of PCA to reduce the dimension of the data as a pre-processing step without losing too much of the important structure\footnote{In contrast to COIL100, on which PCA destroys much of the manifold structure}. We compare the performance of UMAP, FIt-SNE, MulticoreTSNE, and LargeVis on PCA reductions of the Mouse scRNA dataset to varying dimensionalities, and on the original dataset, in Figure \ref{fig:ambient-dim-scaling}. 

While all the implementations tested show a significant increase in runtime with increasing dimension, UMAP is dramatically more efficient for large ambient dimensions, easily scaling to run on the original unreduced dataset. The ability to run manifold learning on raw source data, rather than dimension reduced data that may have lost important manifold structure in the pre-processing, is a significant advantage. This advantage comes from the local connectivity assumption which ensures good topological representation of high dimensional data, particularly with smaller numbers of near neighbors, and the efficiency of the NN-Descent algorithm for approximate nearest neighbor search even in high dimensions.

Since UMAP scales well with ambient dimension the python implementation also supports input in sparse matrix format, allowing scaling to extremely high dimensional data, such as the integer data shown in Figures \ref{fig:umap_3e7_density} and \ref{fig:umap_3e7_value}.

\subsubsection{Scaling with the Number of Samples}\label{data_scaling}

For dataset size performance comparisons we chose to compare UMAP with FIt-SNE \cite{linderman2017efficient}, a version of t-SNE that uses approximate nearest neighbor search and a Fourier interpolation optimisation approach; MulticoreTSNE \cite{Ulyanov2016}, which we believe to be the fastest extant implementation of Barnes-Hut t-SNE; and LargeVis \cite{tang2016visualizing}. It should be noted that FIt-SNE, MulticoreTSNE, and LargeVis are all heavily optimized implementations written in C++. In contrast our UMAP implementation was written in Python --- making use of the numba \cite{Lam:2015:NLP:2833157.2833162} library for performance. MulticoreTSNE and LargeVis were run in single threaded mode to make fair comparisons to our single threaded UMAP implementation.

\begin{figure}[!hptb]
    \centering
    \includegraphics[width=\textwidth]{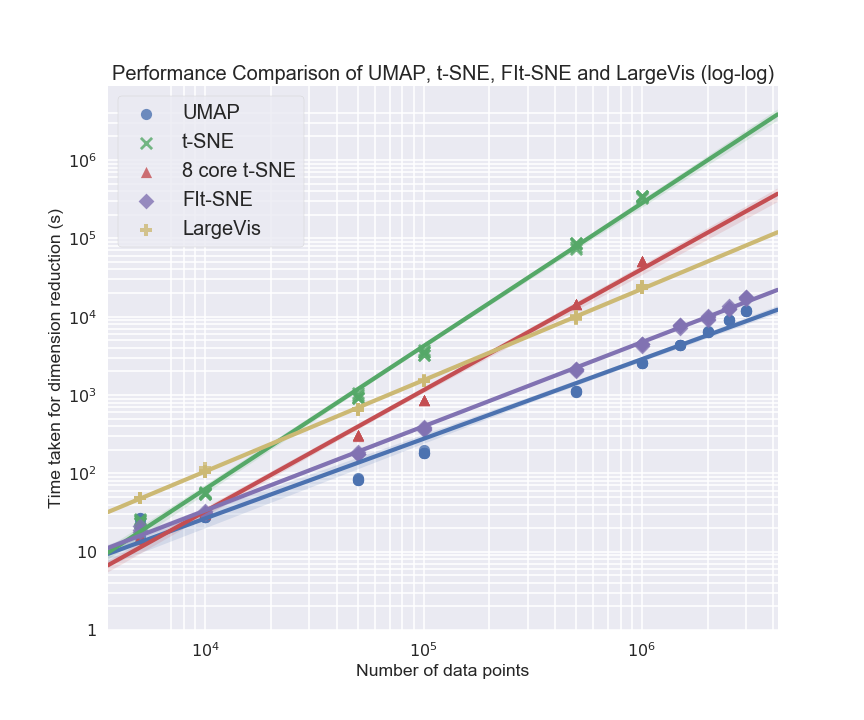}
    \caption{Runtime performance scaling of t-SNE and UMAP on various sized sub-samples of the full Google News dataset. The lower t-SNE line is the wall clock runtime for Multicore t-SNE using 8 cores.}\label{fig:umap_tsne_scaling}
\end{figure}

We benchmarked all four implementations using subsamples of the GoogleNews dataset. The results can be seen in Figure \ref{fig:umap_tsne_scaling}. This demonstrates that UMAP has superior scaling performance in comparison to Barnes-Hut t-SNE, even when Barnes-Hut t-SNE is given multiple cores. Asymptotic scaling of UMAP is comparable to that of FIt-SNE (and LargeVis). On this dataset UMAP demonstrated somewhat faster absolute performance compared to FIt-SNE, and was dramatically faster than LargeVis.

\begin{figure}[!hpbt]
    \centering
    \includegraphics[width=\textwidth]{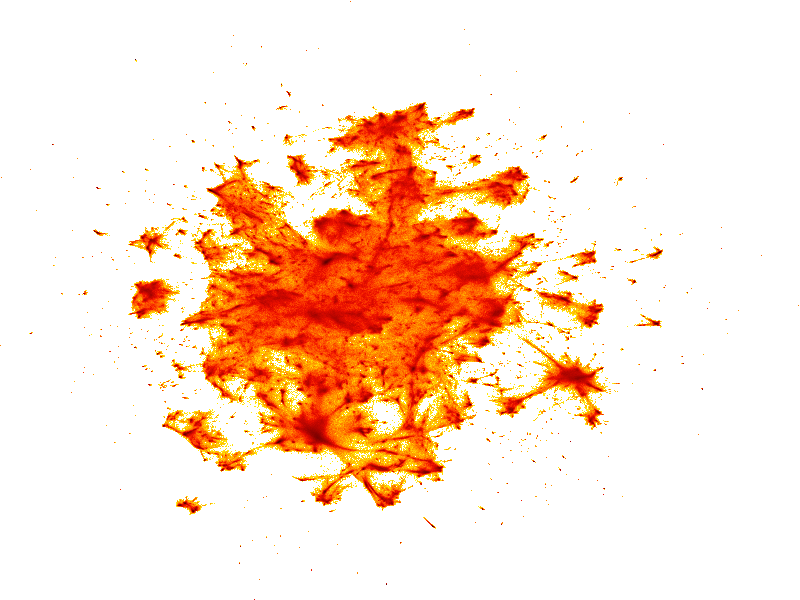}
    \caption{Visualization of the full 3 million word vectors from the GoogleNews dataset as embedded by UMAP.}
    \label{fig:umap_word_vectors}
\end{figure}

The UMAP embedding of the full GoogleNews dataset of 3 million word vectors, as seen in Figure \ref{fig:umap_word_vectors}, was completed in around 200 minutes, as compared with several days required for MulticoreTSNE, even using multiple cores.

\begin{figure}
    \centering
    \includegraphics[width=\textwidth]{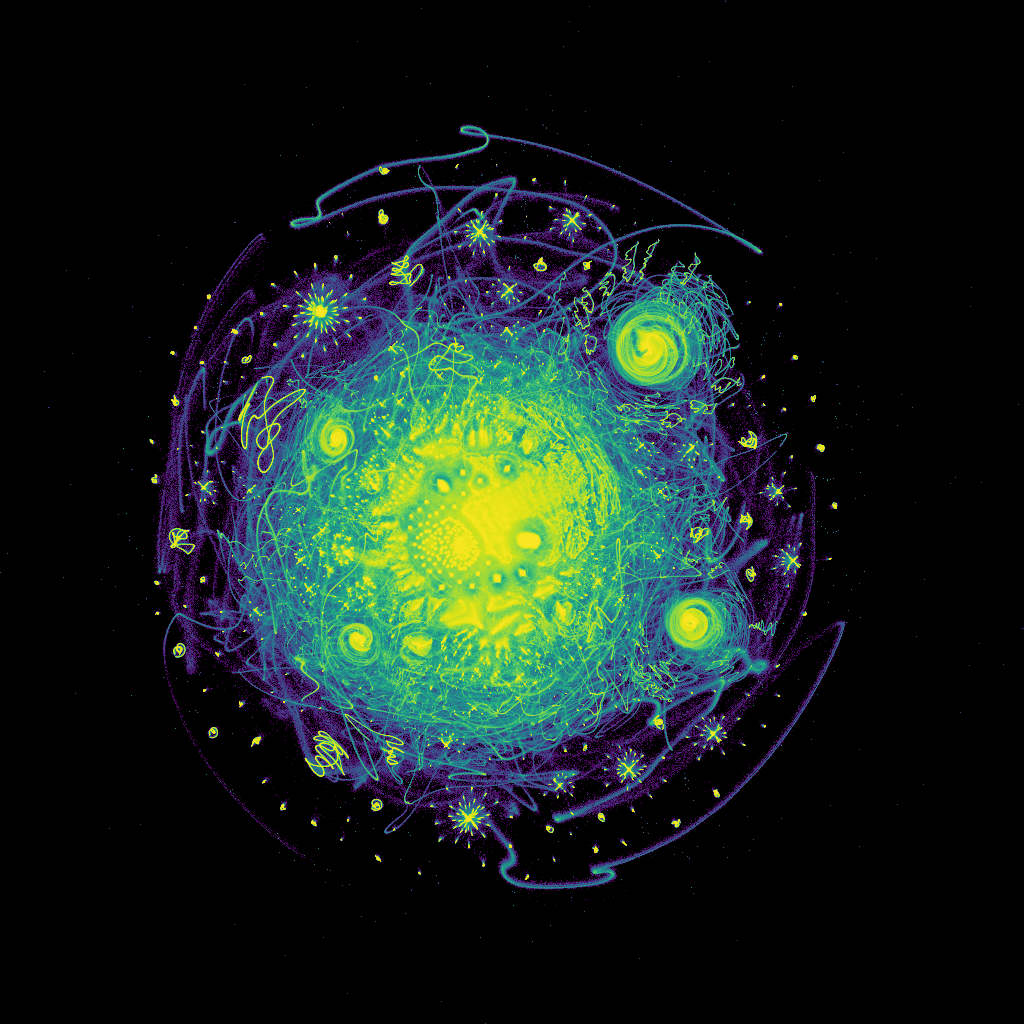}
    \caption{Visualization of 30,000,000 integers as
    represented by binary vectors of prime divisibility, colored by density of points.}
    \label{fig:umap_3e7_density}
\end{figure}

\begin{figure}
    \centering
    \includegraphics[width=\textwidth]{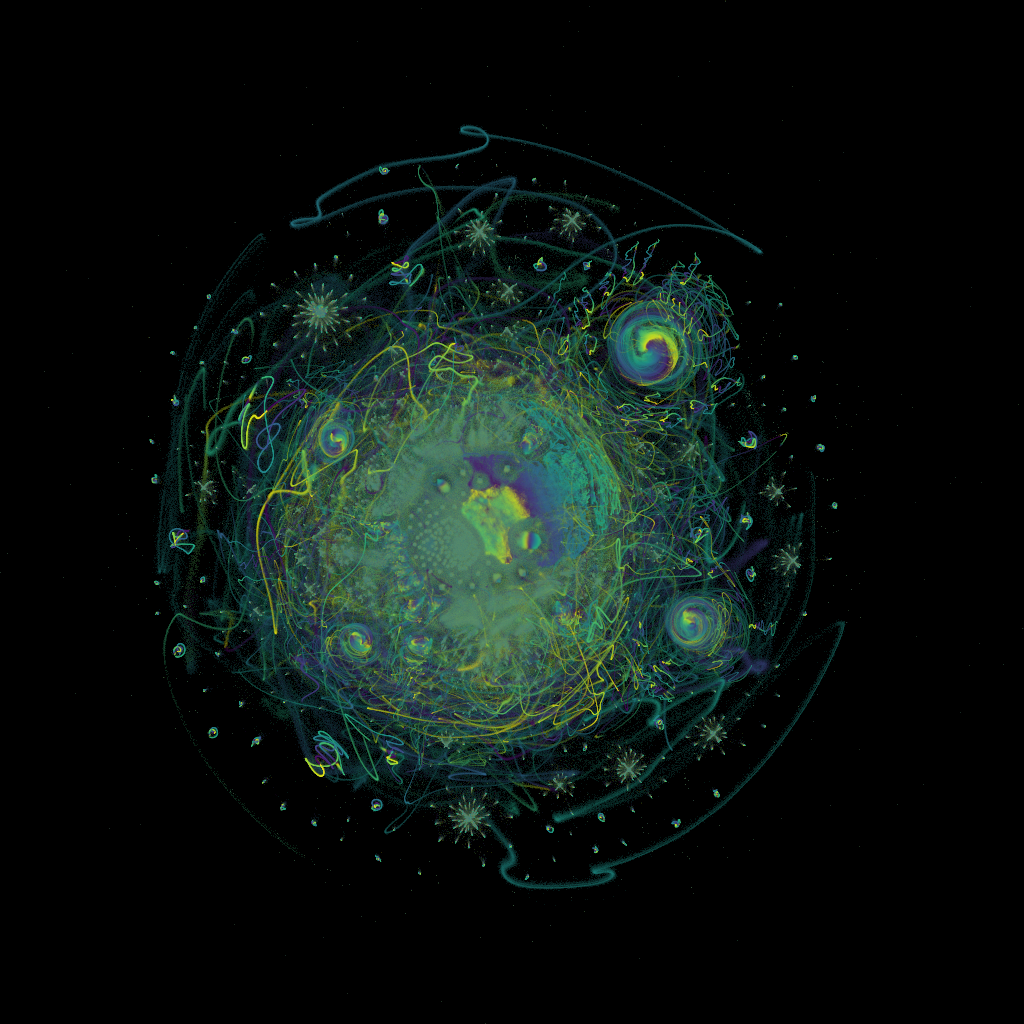}
    \caption{Visualization of 30,000,000 integers as
    represented by binary vectors of prime divisibility, colored by integer value of the point (larger values are green or yellow, smaller values are blue or purple).}
    \label{fig:umap_3e7_value}
\end{figure}

\begin{figure}
    \centering
    \begin{subfigure}{0.45\textwidth}
        \includegraphics[width=\textwidth]{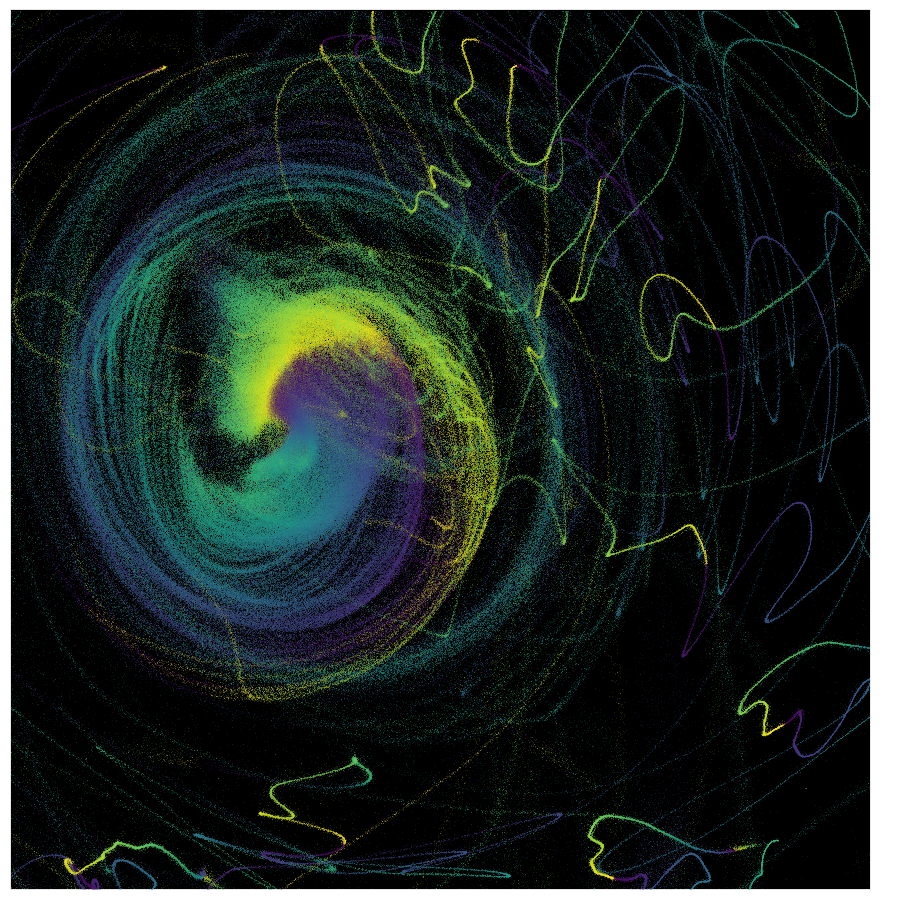}
        \caption{Upper right spiral}
    \end{subfigure}
    \begin{subfigure}{0.45\textwidth}
        \includegraphics[width=\textwidth]{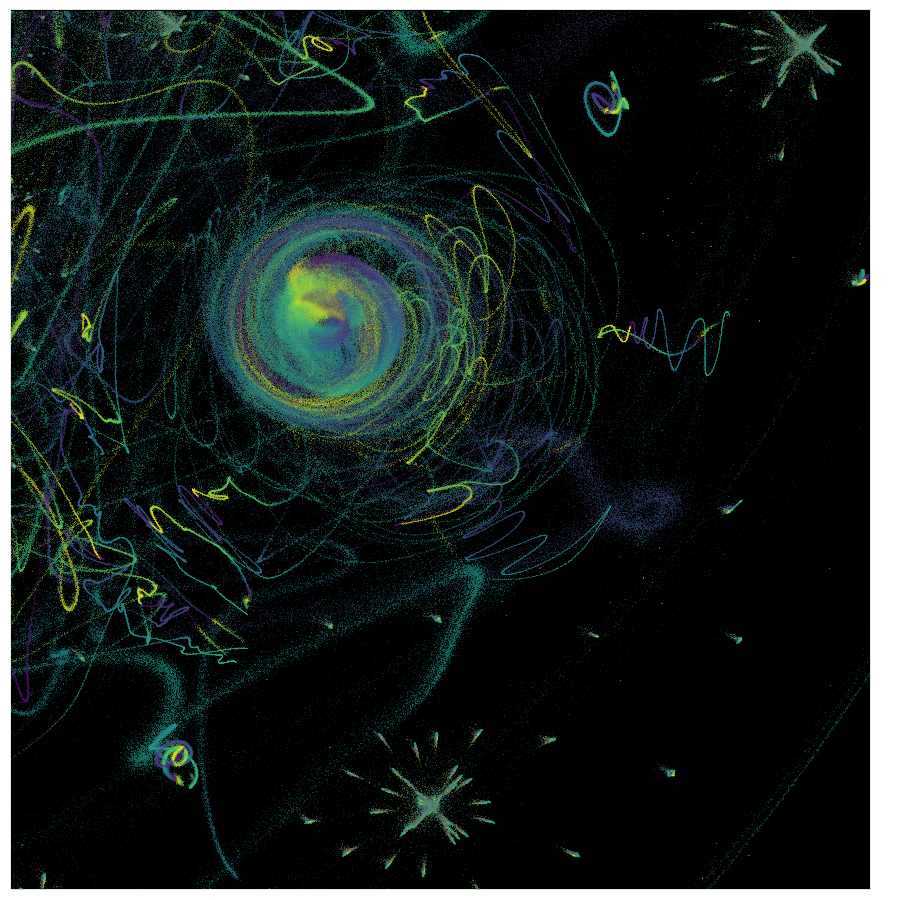}
        \caption{Lower right spiral and starbursts}
    \end{subfigure}
    \begin{subfigure}{0.45\textwidth}
        \includegraphics[width=\textwidth]{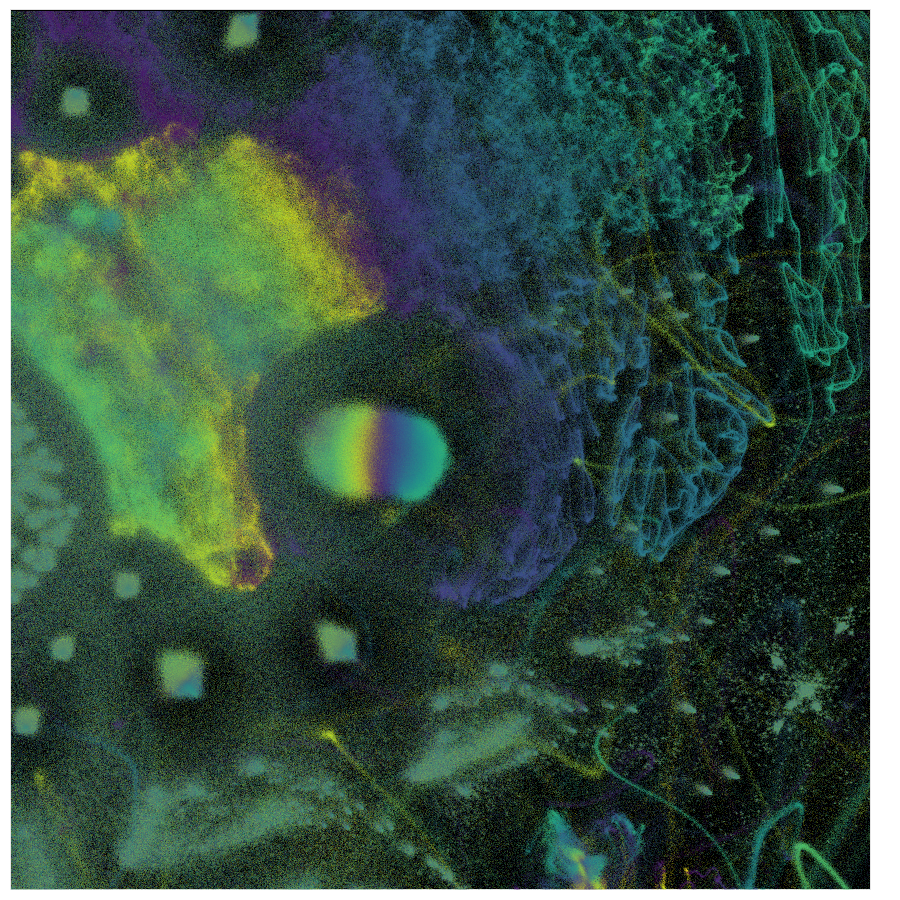}
        \caption{Central cloud}
    \end{subfigure}
    \caption{Zooming in on various regions of
    the integer embedding reveals further layers of
    fine structure have been preserved.}
    \label{fig:umap_3e7_zoom}
\end{figure}

To scale even further we were inspired by the work of John Williamson on embedding integers \cite{williamson_2018}, as represented by (sparse) binary vectors of their prime divisibility. This allows the generation of arbitrarily large, extremely high dimension datasets that still have meaningful structure to be explored. In Figures \ref{fig:umap_3e7_density} and \ref{fig:umap_3e7_value} we show an embedding of 30,000,000 data samples from an ambient space of approximately 1.8 million dimensions. This computation took approximately 2 weeks on a large memory SMP. Note that despite the high ambient dimension, and vast amount of data, UMAP is still able to find and display interesting structure. In Figure \ref{fig:umap_3e7_zoom} we show local regions of the embedding, demonstrating the fine detail structure that was captured.

\section{Weaknesses}\label{weaknesses}

While we believe UMAP to be a very effective algorithm for both visualization and dimension reduction, most algorithms must make trade-offs and UMAP is no exception. In this section we will briefly discuss those areas or use cases where UMAP is less effective, and suggest potential alternatives.

For a number of use cases the interpretability of the reduced dimension results is of critical importance. Similarly to most non-linear dimension reduction techniques (including t-SNE and Isomap), UMAP lacks the strong interpretability of Principal Component Analysis (PCA) and related techniques such a Non-Negative Matrix Factorization (NMF). In particular the dimensions of the UMAP embedding space have no specific meaning, unlike PCA where the dimensions are the directions of greatest variance in the source data. Furthermore, since UMAP is based on the distance between observations rather than the source features, it does not have an equivalent of factor loadings that linear techniques such as PCA, or Factor Analysis can provide. If strong interpretability is critical we therefore recommend linear techniques such as PCA, NMF or pLSA.

One of the core assumptions of UMAP is that there exists manifold structure in the data. Because of this UMAP can tend to find manifold structure within the noise of a dataset -- similar to the way the human mind finds structured constellations among the stars. As more data is sampled the amount of structure evident from noise will tend to decrease and UMAP becomes more robust, however care must be taken with small sample sizes of noisy data, or data with only large scale manifold structure. Detecting when a spurious embedding has occurred is a topic of further research.

UMAP is derived from the axiom that local distance is of more importance than long range distances (similar to techniques like t-SNE and LargeVis). UMAP therefore concerns itself primarily with accurately representing local structure. While we believe that UMAP can capture more global structure than these other techniques, it remains true that if global structure is of primary interest then UMAP may not be the best choice for dimension reduction. Multi-dimensional scaling specifically seeks to preserve the full distance matrix of the data, and as such is a good candidate when all scales of structure are of equal importance. PHATE {\cite{moon2019visualizing}} is a good example of a hybrid approach that begins with local structure information and makes use of MDS to attempt to preserve long scale distances as well. It should be noted that these techniques are more computationally intensive and thus rely on landmarking approaches for scalability. 

It should also be noted that a significant contributor to UMAP's relative global structure preservation is derived from the Laplacian Eigenmaps initialization (which, in turn, followed from the theoretical foundations). This was noted in, for example, {\cite{kobak2019umap}}. The authors of that paper demonstrate that t-SNE, with similar initialization, can perform equivalently to UMAP in a particular measure of global structure preservation. However, the objective function derived for UMAP (cross-entropy) is significantly different from that of t-SNE (KL-divergence), in how it penalizes failures to preserve non-local and global structure, and is also a significant contributor\footnote{The authors would like to thank Nikolay Oskolkov for his article  (\href{https://towardsdatascience.com/tsne-vs-umap-global-structure-4d8045acba17}{tSNE vs. UMAP: Global Structure}) which does an excellent job of highlighting these aspects from an empirical and theoretical basis.}.

It is worth noting that, in combining the local simplicial set structures, pure nearest neighbor structure in the high dimensional space is not explicitly preserved. In particular it introduces so called "reverse-nearest-neighbors" into the classical knn-graph. This, combined with the fact that UMAP is preserving topology rather than pure metric structures, mean that UMAP will not perform as well as some methods on quality measures based on metric structure preservation -- particularly methods, such as MDS -- which are explicitly designed to optimize metric structure preservation.

UMAP attempts to discover a manifold on which your data is uniformly distributed.  If you have strong confidence in the ambient distances of your data you should make use of a technique that explicitly attempts to preserve these distances.  For example if your data consisted of a very loose structure in one area of your ambient space and a very dense structure in another region region UMAP would attempt to put these local areas on an even footing.

Finally, to improve the computational efficiency of the algorithm a number of approximations are made. This can have an impact on the results of UMAP for small (less than 500 samples) dataset sizes. In particular the use of approximate nearest neighbor algorithms, and the negative sampling used in optimization, can result in suboptimal embeddings. For this reason we encourage users to take care with particularly small datasets. A slower but exact implementation of UMAP for small datasets is a future project.

\section{Future Work}\label{future}

Having established both relevant mathematical theory and a concrete implementation, there still remains significant scope for future developments of UMAP. 

A comprehensive empirical study which examines the impact of the various algorithmic components, choices, and hyper-parameters of the algorithm would be beneficial. While the structure and choices of the algorithm presented were derived from our foundational mathematical framework, examining the impacts that these choices have on practical results would be enlightening and a significant contribution to the literature.

As noted in the weaknesses section there is a great deal of uncertainty surrounding the preservation of global structure among the field of manifold learning algorithms. In particular this is hampered by the lack clear objective measures, or even definitions, of global structure preservation. While some metrics exist, they are not comprehensive, and are often specific to various downstream tasks. A systematic study of both metrics of non-local and global structure preservation, and performance of various manifold learning algorithms with respect to them, would be of great benefit. We believe this would aid in better understanding UMAP's success in  various downstream tasks.

Making use of the fuzzy simplicial set representation of data UMAP can potentially be extended to support (semi-)supervised dimension reduction, and dimension reduction for datasets with heterogeneous data types. Each data type (or prediction variables in the supervised case) can be seen as an alternative view of the underlying structure, each with a different associated metric -- for example categorical data may use Jaccard or Dice distance, while ordinal data might use Manhattan distance. Each view and metric can be used to independently generate fuzzy simplicial sets, which can then be intersected together to create a single fuzzy simplicial set for embedding. Extending UMAP to work with mixed data types would vastly increase the range of datasets to which it can be applied. Use cases for (semi-)supervised dimension reduction include semi-supervised clustering, and interactive labelling tools. 

The computational framework established for UMAP allows for the potential development of techniques to add new unseen data points into an existing embedding, and to generate high dimensional representations of arbitrary points in the embedded space. Furthermore, the combination of supervision and the addition of new samples to an existing embedding provides avenues for metric learning. The addition of new samples to an existing embedding would allow UMAP to be used as a feature engineering tool as part of a general machine learning pipeline for either clustering or classification tasks. Pulling points back to the original high dimensional space from the embedded space would potentially allow UMAP to be used as a generative model similar to some use cases for autoencoders. Finally, there are many use cases for metric learning; see \cite{yang2006distance} or \cite{bellet2013survey} for example.

There also remains significant scope to develop techniques to both detect and mitigate against potentially spurious embeddings, particularly for small data cases. The addition of such techniques would make UMAP far more robust as a tool for exploratory data analysis, a common use case when reducing to two dimensions for visualization purposes.

Experimental versions of some of this work are already available in the referenced implementations.

\section{Conclusions}

We have developed a general purpose dimension reduction technique that is grounded in strong mathematical foundations.  The algorithm implementing this technique is demonstrably faster than t-SNE and provides better scaling.  This allows us to generate high quality embeddings of larger data sets than had previously been attainable. The use and effectiveness of UMAP in various scientific fields demonstrates the strength of the algorithm.

\paragraph{Acknowledgements}
The authors would like to thank Colin Weir, Rick Jardine, Brendan Fong, David Spivak and Dmitry Kobak for discussion and useful commentary on various drafts of this paper.
\appendix

\renewcommand\thesection{\Alph{section}}
\setcounter{section}{0}

\section{Proof of Lemma 1}

\setcounter{lem}{0}
\begin{lem}\label{lem:geo-dist}
Let $(\mathcal{M}, g)$ be a Riemannian manifold in an ambient $\mathbb{R}^n$, and let $p \in M$ be a point. If $g$ is locally constant about $p$ in an open neighbourhood $U$ such that $g$ is a constant diagonal matrix in ambient coordinates, then in a ball $B\subseteq U$ centered at $p$ with volume $\frac{\pi^{n/2}}{\Gamma(n/2 + 1)}$ with respect to $g$, the geodesic distance from $p$ to any point $q\in B$ is $\frac{1}{r} d_{\mathbb{R}^n}(p, q)$, where $r$ is the radius of the ball in the ambient space and $d_{\mathbb{R}^n}$ is the existing metric on the ambient space.
\end{lem}

\begin{proof}
Let $x^1, \ldots, x^n$ be the coordinate system for the ambient space. A ball $B$ in $\mathcal{M}$ under Riemannian metric $g$ has volume given by
\[
\int_B \sqrt{\det(g)} dx^1 \wedge \cdots \wedge dx^n .
\]
If $B$ is contained in $U$, then $g$ is constant in $B$ and hence $\sqrt{\det(g)}$ is constant and can be brought outside the integral. Thus, the volume of $B$ is
\[
\sqrt{det(g)}  \int_B dx^1 \wedge ... \wedge dx^n = \sqrt{det(g)} \frac{\pi^{n/2} r^n}{\Gamma(n/2 + 1)},
\]
where $r$ is the radius of the ball in the ambient $\mathbb{R}^n$. If we fix the volume of the ball to be $\frac{\pi^{n/2}}{\Gamma(n/2 + 1)}$ we arrive at the requirement that 
\[
\det(g) = \frac{1}{r^{2n}}.
\]
Now, since $g$ is assumed to be diagonal with constant entries we can solve for $g$ itself as
\begin{equation}\label{eqn:defg}
g_{ij} = \begin{cases}
        \frac{1}{r^2} & \text{ if } i = j,\\[8pt]
        0 & \text{ otherwise}
         \end{cases}.
\end{equation}
The geodesic distance on $\mathcal{M}$ under $g$ from $p$ to $q$ (where $p, q \in B$) is defined as
\[
\inf_{c\in C}\int_a^b \sqrt{g(\dot{c}(t),\dot{c}(t))} dt ,
\]
where $C$ is the class of smooth curves $c$ on $\mathcal{M}$ such that $c(a) = p$ and $c(b) = q$, and $\dot{c}$ denotes the first derivative of $c$ on $\mathcal{M}$. Given that $g$ is as defined in (\ref{eqn:defg}) we see that this can be simplified to
\begin{equation}
    \begin{split}
        &\frac{1}{r}\inf_{c\in C}\int_a^b \langle \sqrt{\dot{c}(t),\dot{c}(t)\rangle} dt\\
        =& \frac{1}{r}\inf_{c\in C}\int_a^b \langle \|\dot{c}(t),\dot{c}(t)\| dt\\
        =& \frac{1}{r} d_{\mathbb{R}^n}(p, q).
    \end{split}
\end{equation}
\end{proof}

\section{Proof that $\FinReal$ and $\FinSing$ are adjoint}

\begin{thm}
The functors $\FinReal:\text{\rm \bf Fin-sFuzz}\to\mathbf{FinEPMet}$ and $\FinSing:\mathbf{FinEPMet}\to \text{\rm \bf Fin-sFuzz}$ form an adjunction with $\FinReal$ the left adjoint and $\FinSing$ the right adjoint.
\end{thm}

\begin{proof}
The adjunction is evident by construction, but can be made more explicit as follows. Define a functor $F:\boldsymbol{\Delta}\times I \to\mathbf{FinEPMet}$ by
\[
F([n], [0,a)) = (\{x_1, x_2, \ldots, x_n\}, d_a),
\]
where
\[
d_a(x_i, x_j) = \begin{cases}
    -\log(a) & \quad\text{if } i \neq j,\\[4pt]
    0 & \quad\text{otherwise}
\end{cases}.
\]
Now $\FinSing$ can be defined in terms of $F$ as 
\[
\FinSing(Y): ([n], [0,a)) \mapsto \hom_{\text{\bf FinEPMet}}(F([n], [0,a)), Y).
\]
where the face maps $d_i$ are given by pre-composition with $Fd^i$, and similarly for degeneracy maps, at any given value of $a$. Furthermore post-composition with $F$ level-wise for each $a$ defines maps of fuzzy simplicial sets making $\FinSing$ a functor.

We now construct $\FinReal$ as the left Kan extension of $F$ along the Yoneda embedding:
\[
\xymatrix{
 & \text{\rm \bf Fin-sFuzz} \ar@{-->}[dr]^{\FinReal} & \\
 \boldsymbol{\Delta} \times I \ar@{^{(}->}[ur]^y \ar[rr]_F & & \mathbf{FinEPMet} 
}
\]
Explicitly this results in a definition of $\FinReal$ at a fuzzy simplicial set $X$ as a colimit:
\[
\FinReal(X) = \colim_{y([n], [0,a)) \to X} F([n]).
\]
Further, it follows from the Yoneda lemma that $\FinReal(\Delta^n_{<a})\cong F([n], [0,a))$, and hence this definition as a left Kan extension agrees with Definition 7, and the definition of $\FinSing$ above agrees with that of Definition 8. To see that $\FinReal$ and $\FinSing$ are adjoint we note that
\begin{equation}\label{eqn:adjoint_iso}
\begin{split}
\hom_{\text{\rm \bf Fin-sFuzz}}(\Delta^n_{<a}, \FinSing(Y)) & \cong \FinSing(Y)^n_{<a}\\
 & = \hom_{\mathbf{FinEPMet}}(F([n], [0,a)), Y)\\
 & \cong \hom_{\mathbf{FinEPMet}}(\FinReal(\Delta^n_{<a}), Y)).
 \end{split}
\end{equation}
The first isomorphism follows from the Yoneda lemma, the equality is by construction, and the final isomorphism follows by another application of the Yoneda lemma. Since every simplicial set can be canonically expressed as a colimit of standard simplices and $\FinReal$ commutes with colimits (as it was defined via a colimit formula), it follows that $\FinReal$ is completely determined by its image on standard simplices. As a result the isomorphism of equation (\ref{eqn:adjoint_iso}) extends to the required isomorphism demonstrating the adjunction.
\end{proof}

\section{From t-SNE to UMAP}\label{compare}

As an aid to implementation of UMAP and to illuminate the algorithmic similarities with t-SNE and LargeVis, here we review the main equations used in those methods, and then present the equivalent UMAP expressions in a notation which may be more familiar to users of those other methods.

In what follows we are concerned with defining similarities between two objects $i$ and $j$ in the high dimensional input space $X$ and low dimensional embedded space $Y$. These are normalized and symmetrized in various ways. In a typical implementation, these pair-wise quantities are stored and manipulated as (potentially sparse) matrices. Quantities with the subscript $ij$ are symmetric, i.e. $v_{ij} = v_{ji}$. Extending the conditional probability notation used in t-SNE, $j \mid i$ indicates an asymmetric similarity, i.e. $v_{j \mid i} \neq v_{i \mid j}$.

t-SNE defines input probabilities in three stages. First, for each pair of points, $i$ and $j$, in $X$, a pair-wise similarity, $v_{ij}$, is calculated, Gaussian with respect to the Euclidean distance between $x_i$ and $x_j$:

\begin{equation}\label{sim_tsne}
    v_{j \mid i} = \exp(-\left\lVert x_i - x_j \right\rVert_2^2 / 2 \sigma_{i}^2)
\end{equation}
where $\sigma_{i}^2$ is the variance of the Gaussian.

Second, the similarities are converted into $N$ conditional probability distributions by normalization:

\begin{equation}\label{prob_tsne}
    p_{j \mid i} = \frac{v_{j \mid i}}{\sum_{k \neq i} v_{k \mid i}} 
\end{equation}
$\sigma_{i}$ is chosen by searching for a value such that the perplexity of the probability distribution $p_{\cdot \mid i}$ matches a user-specified value.

Third, these probability distributions are symmetrized and then further normalized over the entire matrix of values to give a joint probability distribution:

\begin{equation}\label{matrix_norm_tsne}
    p_{ij} = \frac{p_{j \mid i} + p_{i \mid j}}{2N} 
\end{equation}
We note that this is a heuristic definition and not in accordance with standard relationship between conditional and joint probabilities that would be expected under probability semantics usually used to describe t-SNE.

Similarities between pairs of points in the output space $Y$ are defined using a Student t-distribution with one degree of freedom on the squared Euclidean distance:

\begin{equation}\label{weight_tdist}
    w_{ij} = \left(1 + \left\lVert y_i - y_j \right\rVert_2^2 \right) ^ {-1}
\end{equation}
followed by the matrix-wise normalization, to form $q_{ij}$:

\begin{equation}
    q_{ij} = \frac{w_{ij}}{\sum_{k \neq l} w_{kl}} 
\end{equation}
The t-SNE cost is the Kullback-Leibler divergence between the two probability distributions:

\begin{equation}
    C_{t-SNE} = \sum_{i \neq j} p_{ij} \log \frac{p_{ij}}{q_{ij}}
\end{equation}

this can be expanded into constant and non-constant contributions:

\begin{equation}
    C_{t-SNE} = \sum_{i \neq j} p_{ij} \log {p_{ij}}
    -  p_{ij} \log {q_{ij}}
\end{equation}

Because both $p_{ij}$ and $q_{ij}$ require calculations over all pairs of points, improving the efficiency of t-SNE algorithms has involved separate strategies for approximating these quantities. Similarities in the high dimensions are effectively zero outside of the nearest neighbors of each point due to the calibration of the $p_{j \mid i}$ values to reproduce a desired perplexity. Therefore an approximation used in Barnes-Hut t-SNE is to only calculate $v_{j \mid i}$ for $n$ nearest neighbors of $i$, where $n$ is a multiple of the user-selected perplexity and to assume $v_{j \mid i} = 0$ for all other $j$. Because the low dimensional coordinates change with each iteration, a different approach is used to approximate $q_{ij}$. In Barnes-Hut t-SNE and related methods this usually involves grouping together points whose contributions can be approximated as a single point.

A further heuristic algorithm optimization technique employed by t-SNE implementations is the use of \emph{early exaggeration} where, for some number of initial iterations, the $p_{ij}$ are multiplied by some constant greater than 1.0 (usually 12.0). In theoretical analyses of t-SNE such as {\cite{linderman2019clustering}} results are obtained only under an \emph{early exaggeration} regimen with either large constant (of order of the number of samples), or in the limit of infinite exaggeration. Further papers such as {\cite{linderman2017efficient}}, and {\cite{kobak2019art}}, suggest the option of using exaggeration for all iterations rather than just early ones, and demonstrate the utility of this. The effectiveness of these analyses and practical approaches suggests that KL-divergence as a measure between \emph{probability distributions} is not what makes the t-SNE algorithm work, since, under exaggeration, the $p_{ij}$ are manifestly not a probability distribution. This is another example of the probability semantics used to describe t-SNE are primarily descriptive rather than foundational. None the less, t-SNE is highly effective and clearly produces useful results on a very wide variety of tasks.

LargeVis uses a similar approach to Barnes-Hut t-SNE when approximating $p_{ij}$, but further improves efficiency by only requiring approximate nearest neighbors for each point. For the low dimensional coordinates, it abandons normalization of $w_{ij}$ entirely. Rather than use the Kullback-Leibler divergence, it optimizes a likelihood function, and hence is maximized, not minimized:

\begin{equation}
    C_{LV} = \sum_{i \neq j} p_{ij} \log w_{ij} 
+\gamma \sum_{i \neq j} \log \left(1 - w_{ij} \right)
\end{equation}

$p_{ij}$ and $w_{ij}$ are defined as in Barnes-Hut t-SNE (apart from the use of approximate nearest neighbors for $p_{ij}$, and the fact that, in implementation, LargeVis does not normalize the $p_{ij}$ by $N$) and $\gamma$ is a user-chosen positive constant which weights the strength of the the repulsive contributions (second term) relative to the attractive contribution (first term). Note also that the first term resembles the optimizable part of the Kullback-Leibler divergence but using $w_{ij}$ instead of $q_{ij}$. Abandoning calculation of $q_{ij}$ is a crucial change, because the LargeVis cost function is amenable to optimization via stochastic gradient descent.

Ignoring specific definitions of $v_{ij}$ and $w_{ij}$, the UMAP cost function, the cross entropy, is:
\begin{equation}
    C_{UMAP} = \sum_{i \neq j} v_{ij} \log \left( \frac{v_{ij}}{w_{ij}} \right) + 
(1 - v_{ij}) \log \left( \frac{1 - v_{ij}}{1 - w_{ij}} \right)
\end{equation}

Like the Kullback-Leibler divergence, this can be arranged into two constant contributions (those containing $v_{ij}$ only) and two optimizable contributions (containing $w_{ij}$):

\begin{equation}
\begin{aligned}
    C_{UMAP} = 
    \sum_{i \neq j} v_{ij} \log v_{ij}
    + \left(1 - v_{ij}\right) \log \left(1 - v_{ij} \right) \\
    - v_{ij} \log w_{ij}
    - \left(1 - v_{ij}\right) \log \left(1 - w_{ij} \right)
\end{aligned}
\end{equation}

Ignoring the two constant terms, the UMAP cost function has a very similar form to that of LargeVis, but without a $\gamma$ term to weight the repulsive component of the cost function, and without requiring matrix-wise normalization in the high dimensional space. The cost function for UMAP can therefore be optimized (in this case, minimized) with stochastic gradient descent in the same way as LargeVis.

Although the above discussion places UMAP in the same family of methods as t-SNE and LargeVis, it does not use the same definitions for $v_{ij}$ and $w_{ij}$. Using the notation established above, we now provide the equivalent expressions for the UMAP similarities. In the high dimensional space, the similarities $v_{j \mid i}$ are the local fuzzy simplicial set memberships, based on the smooth nearest neighbors distances:

\begin{equation}\label{eq:lfss}
    v_{j \mid i} = \exp[(-d\left(x_i, x_j\right) - \rho_{i}) / \sigma_{i}]
\end{equation}

As with LargeVis, $v_{j \mid i}$ is calculated only for $n$ approximate nearest neighbors and $v_{j \mid i} = 0$ for all other $j$. $d\left(x_i, x_j\right)$ is the distance between $x_i$ and $x_j$, which UMAP does not require to be Euclidean. $\rho_{i}$ is the distance to the nearest neighbor of $i$. $\sigma_{i}$ is the normalizing factor, which is chosen by Algorithm \ref{alg:smooth-knn} and plays a similar role to the perplexity-based calibration of $\sigma_i$ in t-SNE. Calculation of $v_{j \mid i}$ with Equation \ref{eq:lfss} corresponds to Algorithm \ref{alg:fss-construction}. 

Symmetrization is carried out by fuzzy set union using the probabilistic t-conorm and can be expressed as:

\begin{equation}\label{eq:symm}
    v_{ij} = \left(v_{j \mid i} + v_{i \mid j}\right) - v_{j \mid i} v_{i \mid j}
\end{equation}
Equation \ref{eq:symm} corresponds to forming $\text{top-rep}$ in Algorithm \ref{alg:umap}. Unlike t-SNE, further normalization is not carried out.

The low dimensional similarities are given by:
\begin{equation}
w_{ij} = \left(1 + a \left\lVert y_i - y_j \right\rVert _2^{2b}\right) ^ {-1}
\end{equation}
where $a$ and $b$ are user-defined positive values. The procedure for finding them is given in Definition \ref{defn:nu_approx}. Use of this procedure with the default values in the UMAP implementation results in $a \approx 1.929$ and $b \approx 0.7915$. Setting $a = 1$ and $b = 1$ results in the Student t-distribution used in t-SNE.

\bibliography{references}{}
\bibliographystyle{plain}

\end{document}